%% file: main.tex
\documentclass[runningheads]{llncs}

\usepackage{eccv}

\usepackage{eccvabbrv}
\usepackage{cite}

\usepackage{graphicx}
\usepackage{arydshln}
\usepackage{booktabs}
\usepackage{multirow}
\usepackage{moresize}
\usepackage[noend]{algpseudocode}
\usepackage{algorithm}
\usepackage{listings}
\usepackage[accsupp]{axessibility}  % Improves PDF readability for those with disabilities.

\newcommand{\first}[1]{\bf{#1}}
\newcommand{\second}[1]{\underline{#1}}

\usepackage{hyperref}

\usepackage{orcidlink}

\definecolor{codegreen}{rgb}{0,0.6,0}
\definecolor{codegray}{rgb}{0.5,0.5,0.5}
\definecolor{codepurple}{rgb}{0.58,0,0.82}
\definecolor{backcolour}{rgb}{0.95,0.95,0.92}
\lstdefinestyle{mystyle}{
    backgroundcolor=\color{backcolour},   
    commentstyle=\color{codegreen},
    keywordstyle=\color{magenta},
    numberstyle=\tiny\color{codegray},
    stringstyle=\color{codepurple},
    basicstyle=\ttfamily\footnotesize,
    breakatwhitespace=false,         
    breaklines=true,                 
    captionpos=b,                    
    keepspaces=true,                 
    numbers=left,                    
    numbersep=5pt,                  
    showspaces=false,                
    showstringspaces=false,
    showtabs=false,                  
    tabsize=2
}

\lstdefinestyle{myPython}{
 	language = Python,
 	breaklines = true,
 	breakindent = 10pt,
 	basicstyle = \ttfamily\scriptsize,
 	commentstyle = {\itshape \color[cmyk]{1,0.4,1,0}},
 	classoffset = 1,
 	stringstyle = {\ttfamily \color[rgb]{0,0,1}},
 	numberstyle = \tiny,
 	tabsize = 4,
 	captionpos = t,
        alsoletter= [1]{.},
        morekeywords = [1]{torch.Tensor, torch},
        keywordstyle = [1]{\bfseries \color[cmyk]{0.7,0,0,0.2}},
}

\usepackage{orcidlink}
\begin{document}

% ---------------------------------------------------------------
\title{Object-Aware Query Perturbation \\ for Cross-Modal Image-Text Retrieval}

\author{Naoya Sogi \and
Takashi Shibata \and
Makoto Terao
}

\authorrunning{N.~Sogi et al.}

\institute{Visual Intelligence Research Laboratories, NEC Corporation,
Kanagawa, Japan \\
\email{naoya-sogi@nec.com, t.shibata@ieee.org, m-terao@nec.com}\\
}

\maketitle
\begin{abstract}
The pre-trained vision and language (V\&L) models have substantially improved the performance of cross-modal image-text retrieval. In general, however, V\&L models have limited retrieval performance for small objects because of the rough alignment between words and the small objects in the image. In contrast, it is known that human cognition is object-centric, and we pay more attention to important objects, even if they are small. To bridge this gap between the human cognition and the V\&L model's capability, we propose a cross-modal image-text retrieval framework based on ``object-aware query perturbation.'' The proposed method generates a key feature subspace of the detected objects and perturbs the corresponding queries using this subspace to improve the object awareness in the image. In our proposed method, object-aware cross-modal image-text retrieval is possible while keeping the rich expressive power and retrieval performance of existing V\&L models without additional fine-tuning. Comprehensive experiments on four public datasets show that our method outperforms conventional algorithms.
Our code is publicly available at \url{https://github.com/NEC-N-SOGI/query-perturbation}

\keywords{cross-modal retrieval \and vision and language \and object centric}
\end{abstract}

\section{Introduction}

% FIXME;
% \footnote[0]{doi}
\begin{figure}[t]
    \centering
    \includegraphics[width=0.75\columnwidth]{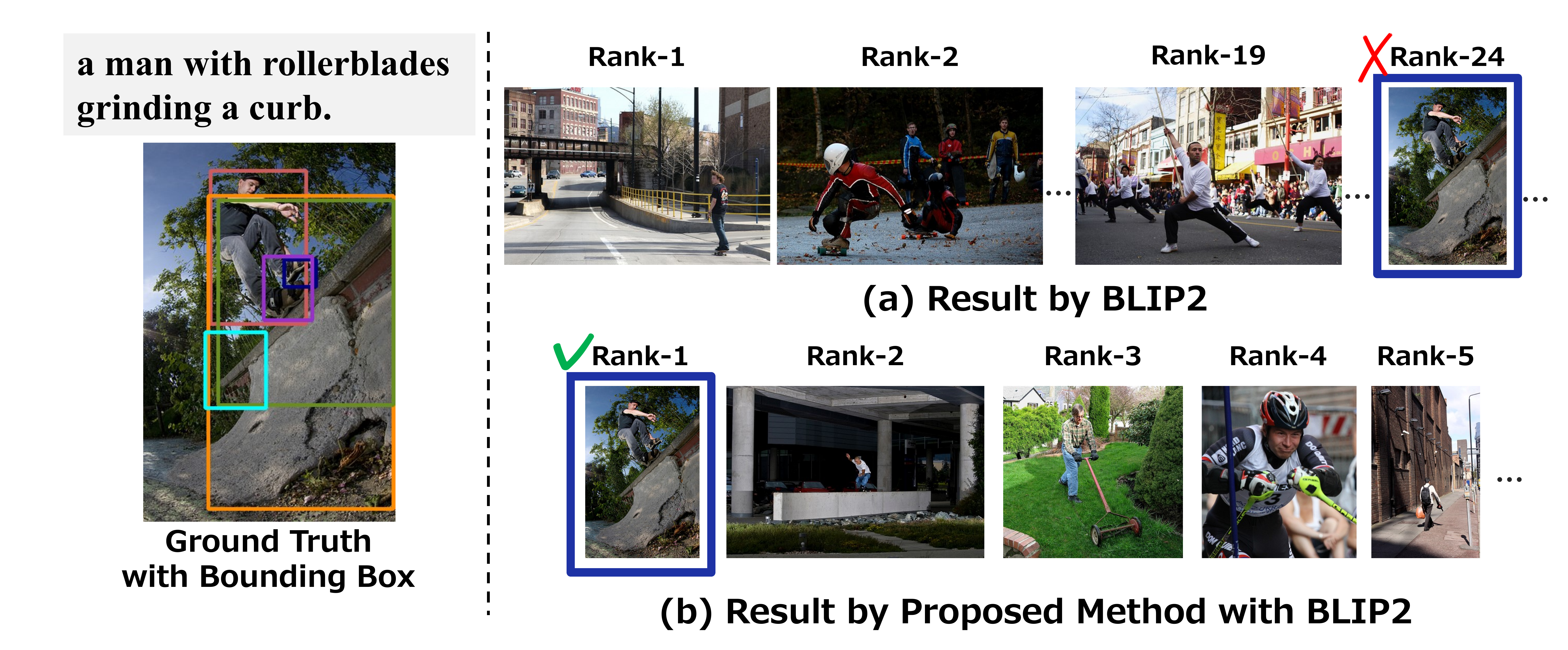}
    %\includegraphics[width=0.8\columnwidth]{figs/eccv-fig1-r4.png}
    %\vspace{-0.2cm}
    \caption{Example results by our method with BLIP2~\cite{BLIP2BootstrappingLanguageImagea}. In BLIP2, the matching between the target objects and the input text is weak because the objects are small, resulting in incorrect retrieval results.}
    \label{fig:1}
    %\vspace{-0.2cm}
\end{figure}

Cross-modal image-text retrieval is one of the mainstream tasks in pattern recognition~\cite{ComparativeAnalysisCrossmodal,ImagetextRetrievalSurvey} and has various applications including e-commerce~\cite{goenka2022fashionvlp,KnowledgeawareMultimodalDialogue} and video surveillance~\cite{jiang2023cross}.
Recent pre-trained vision-and-language (V\&L) models~\cite{clip,blip,albef,chen2020uniter,beit,yu2022coca} have caused a paradigm shift.
Those pre-trained models substantially outperform legacy cross-modal image-text retrieval by leveraging massive amounts of training data while equipping advantages such as zero-shotness and generalizability.

Nevertheless, those V\&L models are not any panacea; those V\&L models have limited performance for small objects due to the rough alignment between text and the fine-grained localization of these small targets in the image.
An example of retrieval results obtained by a sophisticated V\&L model, BLIP2~\cite{BLIP2BootstrappingLanguageImagea}, on the Flickr 30K~\cite{flickrentitiesijcv} dataset is shown in Fig.~\ref{fig:1}(a).
The matching between the target objects and the input query text is weak because the objects, e.g., the person and the rollerblades, are small, resulting in incorrect retrieval results.
Although the drawback of the retrieval performance degradation related to those small objects is critical in actual applications, it has been hidden behind the overwhelming performance gains of the recent pre-trained V\&L models on the public benchmark datasets.

In contrast, humans can effectively understand visual scenes by an ability that lies in their object-centered (or compositional) perception~\cite{lake2017building,yuan2023compositional}. 
Owing to this human object-centered perception, the human visual function is highly robust to the size of the target object.
For example, objects critical to understanding a scene, e.g., a small rescue caller in an image of a disaster scene, will be gazed at regardless of the target object's size.

The lack of such object-awareness in V\&L models is a major issue, especially for human-centered vision tasks, e.g., image retrieval.
Although legacy image retrieval algorithms using object detection have also been proposed~\cite{ji2021step,lee2018stacked}, these methods cannot inherit the strengths of recent pre-trained V\&L models.
There is a strong demand for a general framework that bridges the gap between human perception and the V\&L models while inheriting the potential capabilities of pre-trained V\&L models including zero-shotness and absolute high performance. 

This paper proposes an object-aware query-perturbation for cross-modal retrieval as a solution to the above demand.
Our Query-Perturbation (Q-Perturbation) increases object awareness of V\&L models by focusing on object information of interest even when the objects in an image are relatively small.
An example of retrieval results by the proposed method is shown in Fig.~\ref{fig:1}(b).
In contrast to the existing methods, our retrieval framework, i.e., a V\&L model with Q-Perturbation, can perform accurate retrieval even for images that capture small objects.
The core mechanism of Q-Perturbation is to enhance queries with keys corresponding to object regions, at cross-attention modules in a V\&L model.

Naively enhancing queries in the existing V\&L model breaks the original weights, resulting in poor performance.
Our query perturbation prevents weight breaking by enhancing queries within a subspace constructed using keys corresponding to target objects.
That is, queries are first decomposed by subspaces representing object information and then enhanced using the decomposed information, which is object information retained in original queries.
This process naturally selects queries to be enhanced as the process uses only object information in original queries, i.e., a query is not enhanced if it does not have object information.
As a result, our Query Perturbation improves the object-awareness of a V\&L model while inheriting the impressive performance of a V\&L model.

The proposed method is applicable to a variety of V\&L models and can avoid increased computational cost due to data updates and catastrophic forgetting due to re-training because the proposed method is training-free and easy to implement.
Comprehensive experiments on public datasets demonstrate the effectiveness of the proposed method.
% FIXME; url to repository

%\if0
The contributions of this paper are including:
1) We propose an object-aware query perturbation (Q-Perturbation) for cross-modal image-text retrieval. 
2) We construct the object-aware retrieval framework by plugging Q-Perturbation into state-of-the-art V\&L models, e.g.,  BLIP2~\cite{BLIP2BootstrappingLanguageImagea}, COCA~\cite{yu2022coca}, and InternVL~\cite{chen2023internvl}. 
3) Comprehensive experiments on public data demonstrate the effectiveness of the proposed method. In addition, we propose a new metric that mitigates the dataset bias regarding object size.
%\fi

\section{Related works}
\textbf{Cross-Modal Image-Text Retrieval.}
The study of cross-modal image-text retrieval is a fundamental task in vision, and many existing methods exist~\cite{chen2021learning,VSEImprovingVisualSemantic,gu2018look,ji2019saliency,wang2018learning,yan2021discrete,huang2017instance,lee2018stacked,liu2019focus,wang2019camp,wei2020multi,zhang2020context,wang2020consensus,StepWiseHierarchicalAlignment,ViLEMVisualLanguageError,LearningHierarchicalSemantic,IMRAMIterativeMatching,zhang2018deep}. 
% \footnote{A more detailed survey is these~\cite{ComparativeAnalysisCrossmodal,ImagetextRetrievalSurvey}.}
A standard approach acquires a text-language common space from image and text datasets prepared as training data in advance~\cite{chun2021probabilistic,engilberge2018finding,VSEImprovingVisualSemantic,karpathy2014deep,song2019polysemous,thomas2020preserving,li2019visual}. 
To acquire an accurate image-text common space, several approaches have been introduced to improve the loss function and distance space, such as metric learning~\cite{VSEImprovingVisualSemantic,zhang2018deep,zheng2019towards} and probabilistic distribution representation~\cite{chun2021probabilistic,ImprovingCrossModalRetrieval,MultilateralSemanticRelations}. 
Various extensions have been proposed for fine-grained retrieval~\cite{chen2020interclass,lee2018stacked,liu2019focus,zhang2022negative,wei2020universal} by introducing object detection~\cite{lee2018stacked,ji2021step}, graph-based relationships between objects~\cite{diao2021similarity,liu2020graph,zhang2022show}, re-weighting strategy~\cite{wei2021meta,chen2020interclass,wei2020universal}, and the attention mechanism~\cite{zhang2020context,IMRAMIterativeMatching}. 
These existing approaches for fine-grained retrieval suggest that object awareness is an essential cue for locally detailed cross-modal retrieval. 
This paper focuses on object awareness for the pre-trained V\&L models~\cite{PiTLCrossmodalRetrieval,albef,blip,BLIP2BootstrappingLanguageImagea,ALIGN,GivingTextMore,ProbVLMProbabilisticAdapter}. 
We propose a simple-yet-effective framework that can efficiently improve the performance of image-text retrieval for an image containing small objects that are semantically important.

%\subsection{Pretrained Vision \& Language Model}
\vspace{5pt} \noindent
\textbf{Pre-trained Vision \& Language Model.}
In recent years, cross-modal image-text retrieval using the V\&L model has been proposed as a new paradigm~\cite{PiTLCrossmodalRetrieval,albef,blip,BLIP2BootstrappingLanguageImagea,ALIGN,GivingTextMore,ProbVLMProbabilisticAdapter}. 
Vision language pre-training, such as CLIP~\cite{clip}, trains vision language alignments from large numbers of image-text pairs through a self-supervised task. 
Before this paradigm, the existing image-text retrieval methods mainly focused on training algorithms using medium-sized datasets such as Flicker 30K and COCO.  
In contrast, the recent cross-modal image-text retrieval using the pre-trained V\&L models outperforms those legacy cross-modal image-text retrieval methods, achieving high zero-shot performance on diverse datasets and enabling open vocabulary retrieval. 
In particular, the recently proposed BLIP2~\cite{BLIP2BootstrappingLanguageImagea} has overwhelming performance in cross-modal image-text retrieval. 
However, it has recently been pointed out that there is a weakness in the localization for the V\&L models such as CLIP, and several simple improvements have been proposed~\cite{zhong2022regionclip,shtedritski2023does}. 
As described later, this weakness has also been shared in cross-modal image-text retrieval. 
The proposed method is a novel framework that can overcome this weakness through cross-modal image-text retrieval while taking advantage of the potential capabilities of the existing V\&L models.

\begin{figure}[t]
    \centering
    \includegraphics[width=0.95\columnwidth]{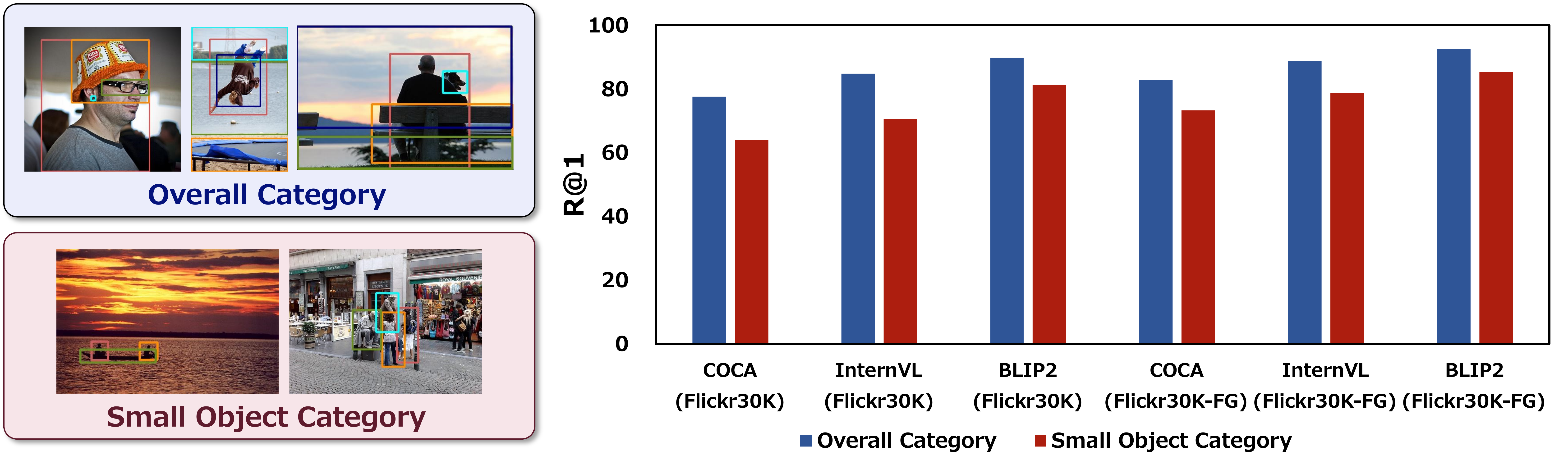}\\
    \caption{Performance degradation induced by small objects.} 
    \label{fig:analysis}
\end{figure}
\section{Performance Degradation Induced by Small Objects}
We discuss the performance degradation induced by small objects in a target image. 
We compared the overall performance of text-image retrieval (called overall category) on Flickr-30K~\cite{flickrentitiesijcv} and Flickr-FG~\cite{RethinkingBenchmarksCrossmodal} with the performance on a dataset consisting of images with only small detected objects (called small object category).
Specifically, we selected images for the small object category, where the ratio of the largest detected object rectangle's area to the entire image's area is less than 10\%.
Figure~\ref{fig:analysis} shows the comparison results.
Here, we used Recall@1 as the evaluation metric.
It can be seen that as the area of the object detection rectangle becomes relatively smaller, the retrieval becomes more difficult, and the retrieval performance degrades.
 
The degradation is observed not only in Flickr-30K but also in Flickr-FG, where more detailed captions are annotated. 
Interestingly, we also find that this is a common drawback in the recent pre-trained V\&L models~\cite{BLIP2BootstrappingLanguageImagea,yu2022coca,chen2023internvl}. 
This drawback is underestimated in the standard evaluation metric, Recall@K, because the number of images belonging to the small object category is small.
For example, the number of images in the small object category accounts for about 1.5\% of all images in Flickr-FG and Flickr-30K.
We introduce an object-aware query perturbation to improve the poor performance of images with such small objects. 
Furthermore, we also discuss the effectiveness of the proposed method using an evaluation metric that accounts for this data bias regarding the object size in Sec~\ref{sec:exp}.

\section{Method}
\begin{figure}[t]
    \centering
    \includegraphics[width=0.97\columnwidth]{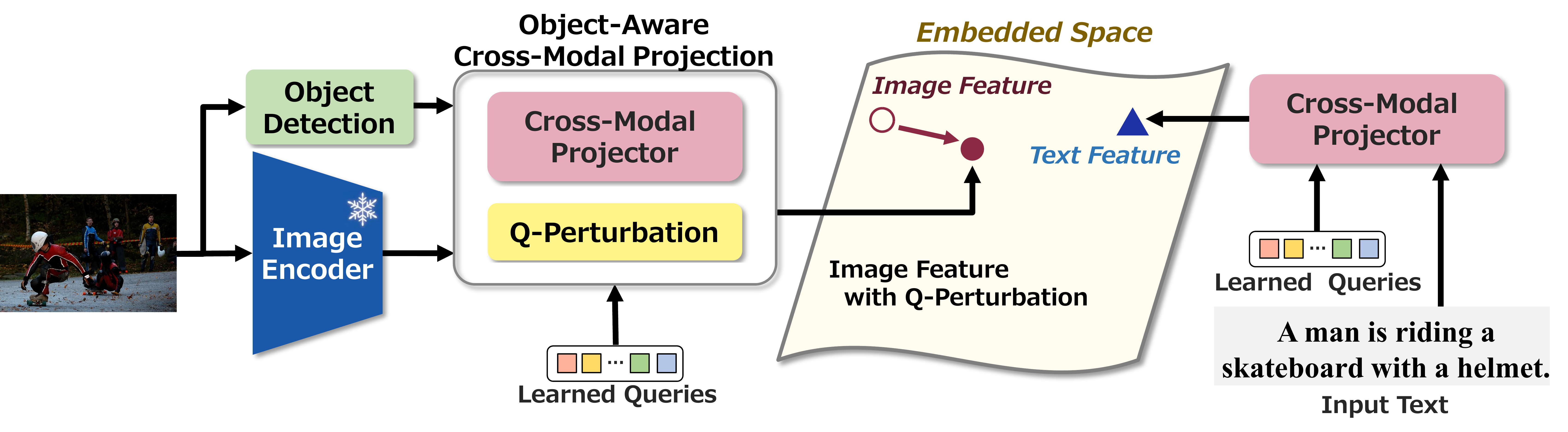}\\
    \caption{Overview of the proposed framework. The proposed framework constructs an object-aware cross-modal projector by incorporating localization cues from object detection into the existing cross-modal projector.}
    \label{fig:overview}
\end{figure}
We describe the overview of our framework and key idea of the query perturbation (Q-Perturbation), then describe the details of the Q-Perturbations for the Q-Former module in BLIP2~\cite{BLIP2BootstrappingLanguageImagea}. Finally, we explain an extension of our Q-Perturbations to other V\&L models~\cite{yu2022coca,chen2023internvl}.

\subsection{Overview}
Our proposed method aims at improving the retrieval performance for an image containing small objects by extending the existing V\&L models while inheriting the high expressiveness of these models.
In general, V\&L models contain a cross-modal projection module that aligns language features with image features.
For example, BLIP2 introduces Q-Former architecture with a transformer structure as a cross-modal projector that combines image and text features. 

An overview of our proposed framework is shown in Fig.~\ref{fig:overview}. 
The proposed framework is also an architecture that leverages cross-modal projectors such as Q-Former~\cite{BLIP2BootstrappingLanguageImagea} and QLLaMA\cite{chen2023internvl}. 
The input text, a retrieval query, is similar to a standard cross-modal retrieval, and feature representations of texts are generated using a text encoder. 
The proposed framework constructs an object-aware cross-modal projector by incorporating localization cues obtained from object detection into the existing cross-modal projector, in addition to image features obtained from existing image encoders. 
The key is how to incorporate the localization cues into existing cross-modal projection modules.
To do this, the proposed method introduces an object-aware query perturbation, called \textit{Q-Perturbation}, that adaptively adjusts the query according to the size of the detected objects and bounding boxes in the image.

\subsection{Basic Idea: Object-Aware Query Perturbation}
A standard approach to constructing a cross-modal projector in a transformer-based module is to introduce cross-attention.
For example, in BLIP2, cross-attention is introduced in the Q-Former to integrate image features with queries, including learned queries and text tokens.

Our goal is to incorporate object localization cues from the bounding boxes into cross-modal projection modules with minimal modification while taking advantage of the highly expressive power of the existing V\&L models. To this end, the following must be satisfied:

\noindent
- \textbf{Inheritability}: The proposed method must be object-aware cross-modal projection without significantly destroying the weights and structures already learned, in order to maximize the potential of the existing V\&L models. 

\noindent
- \textbf{Flexibility}: The proposed method must be scalable and flexible regarding the size and number of detected objects.

\begin{figure}[t]
    \centering
    \begin{minipage}[b]{.43\hsize}
        \includegraphics[width=0.85\columnwidth]{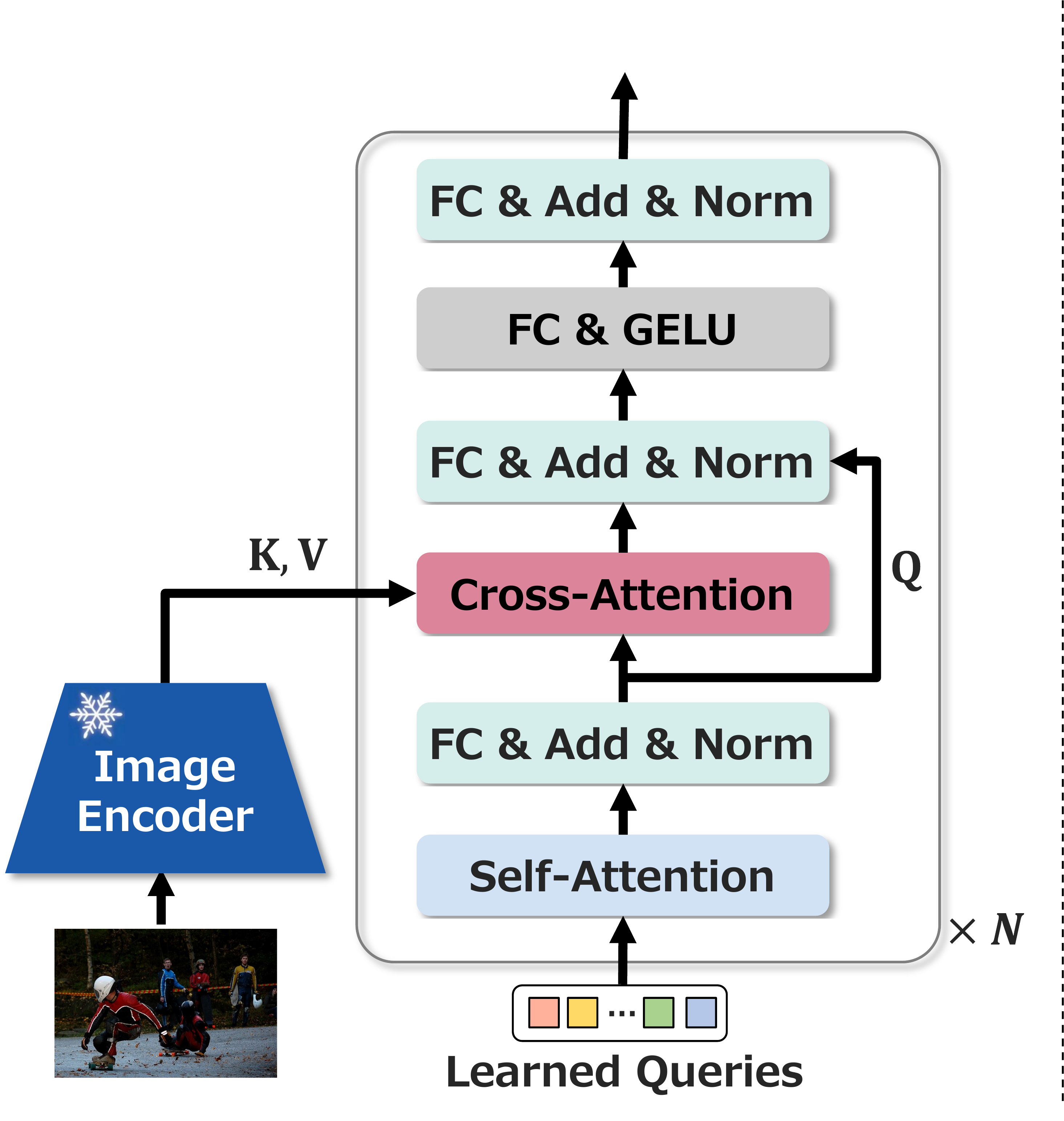}
        \subcaption{Q-Former.}
    \end{minipage}
    \begin{minipage}[b]{.43\hsize}
        \includegraphics[width=0.97\columnwidth]{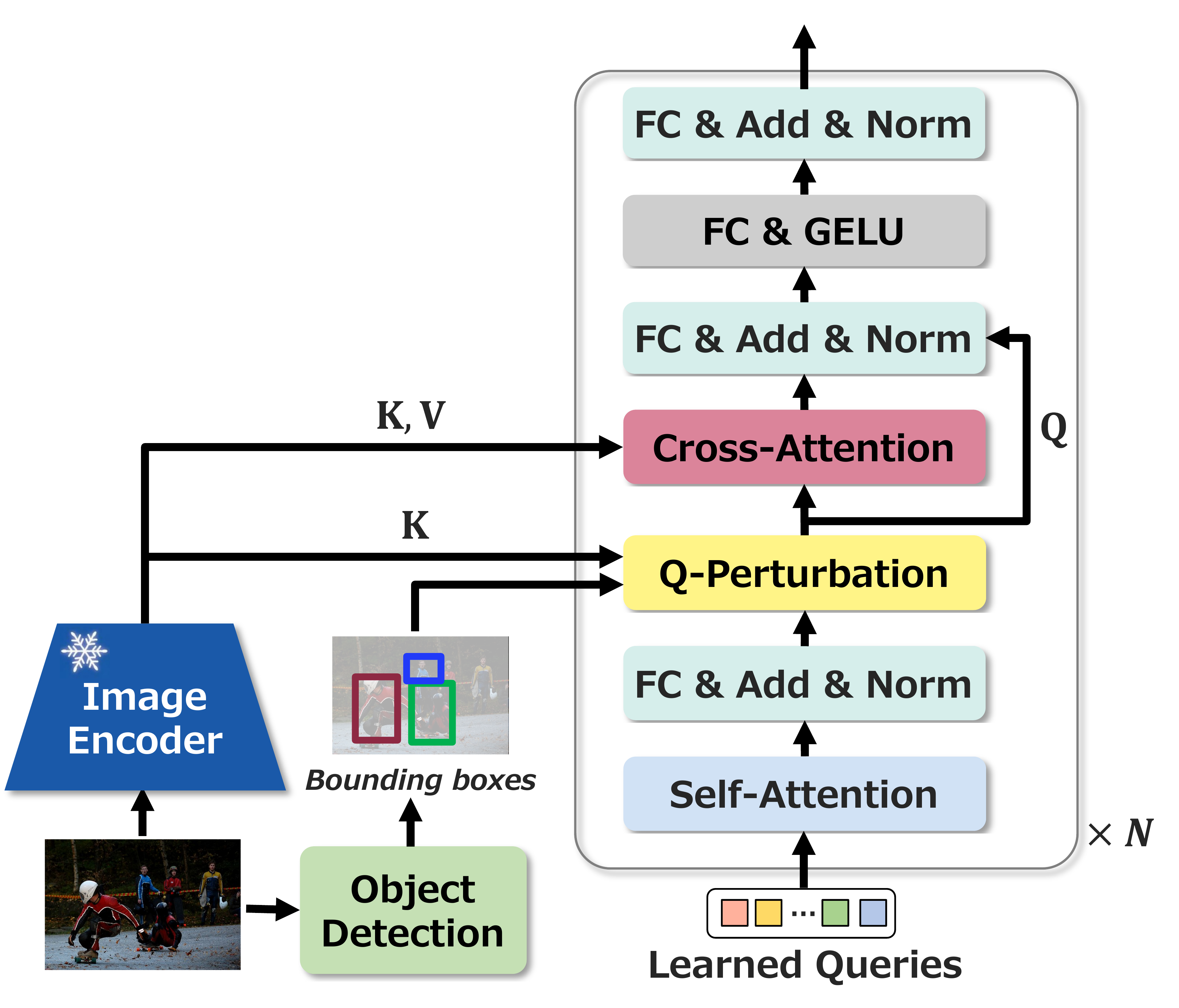}
        \subcaption{Q-Former with the proposed Q-Perturbation.}
    \end{minipage}
    \caption{Overview of our object-aware mechanism with Q-Perturbation module. Our framework can incorporate the object localization cues from the bounding boxes into a cross-modal projection module, e.g., Q-Former, with minimal modification.}
    \label{fig;q-perturb-overview}
\end{figure}

In the proposed method, as shown in Fig.~\ref{fig;q-perturb-overview}, the Q-Perturbation is used to perturb the already obtained query to emphasize the object region features using object localization, i.e., bounding box.
An object-aware cross-modal projection module can be implemented with minimal modifications by plugging the proposed Q-Perturbation module just before the cross-attention module. 
In the following, we first describe our Q-Perturbation module for the single object case and then extend it to multiple objects.

\subsection{Q-Perturbation Module for Single Objects}
\begin{figure}[t]
    \centering
    \includegraphics[width=1\columnwidth]{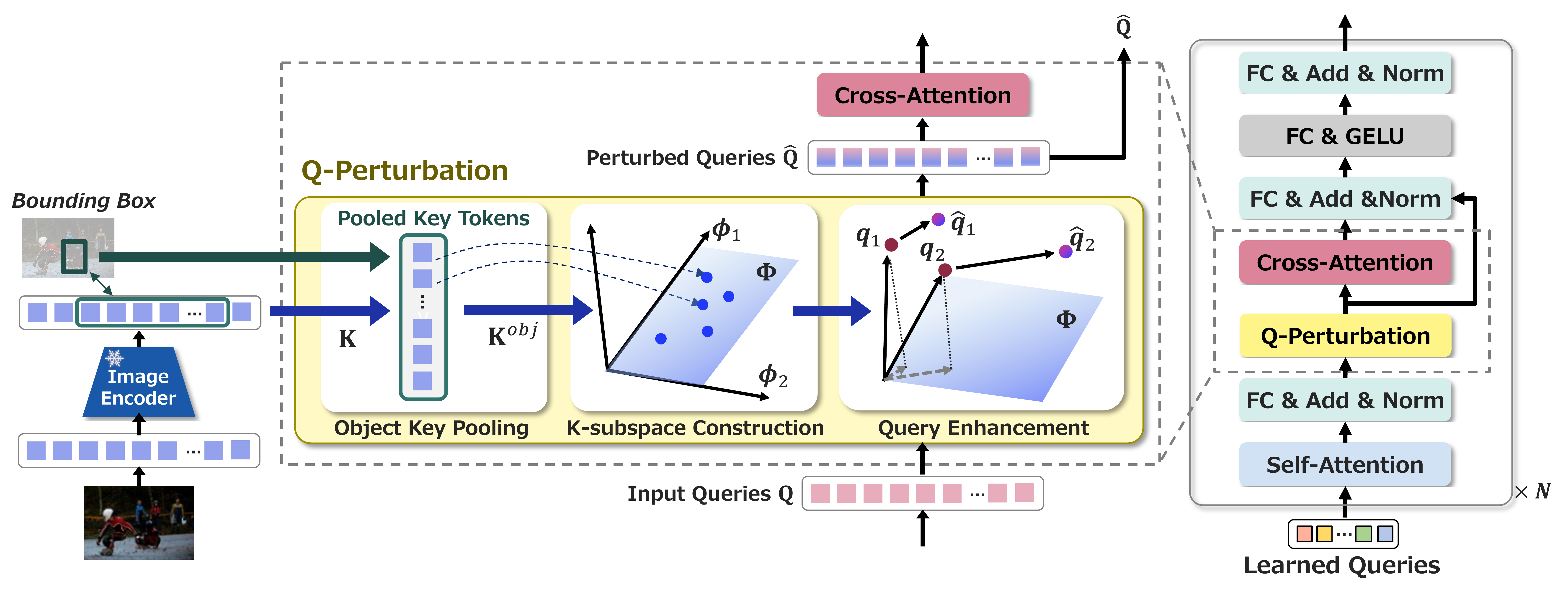}\\
    \caption{Overview of our Q-Perturbation. The Q-Perturbation consists of three components: Object Key Pooling, K-subspace Construction, and Query Enhancement.}
    \label{fig:q-perturb-detail}
\end{figure}

The proposed Q-Perturbation consists of three components: 1) Object Key Pooling, 2) K-Subspace Construction, and 3) Query Enhancement as shown in Fig.~\ref{fig:q-perturb-detail}.
Let $ {\bf{Q}} = \{ {\bf{q}}_{i} \}$, ${\bf{K}} =  \{ {\bf{k}}_{l} \}$, and ${\bf{V}} =  \{ {\bf{v}}_{l} \}$ be the set of queries before the cross-attention, and the set of the keys and values from the image encoder, respectively.
Here, $i$ and $l$ are subscripts to distinguish each token.

\vspace{3pt} \noindent \textbf{1) Object Key Pooling.}
First, we select image tokens that overlap the detected bounding box obtained by object detection, as shown in Fig.~\ref{fig:q-perturb-detail}.
Our pooling step is in the same manner as ROI pooling in two-stage object detection~\cite{ren2015faster}.
In the following, those selected image tokens are called object image tokens, and the pooled set is denoted as ${\bf{K}}^{obj} = \{ {\bf{k}}^{obj}_{j} \}$, where $j$ is subscripts to distinguish each token.
Note that, if there are multiple objects in an image, we perform the object key pooling for each detected object.

\vspace{3pt} \noindent \textbf{2) K-Subspace Construction.}
Next, the object-aware key subspace, called \textit{K-subspace}, is generated from the pooled object image tokens ${{\bf{k}}^{obj}_j}$ for each object using Principal Component Analysis (for more detail, see supplementary material A.2).
The K-subspace for each object is denoted by ${\bf{\Phi}}=[{\bf{\phi}}_{1}, {\bf{\phi}}_{2}, \cdots, {\bf{\phi}}_{p}, \cdots ]$, where ${\bf{\phi}}_{p}$ is the $p$-th basis vector of the K-subspace.
The K-subspace represents essential information about the corresponding object.

\vspace{3pt} \noindent \textbf{3) Query Enhancement.}
Finally, the set of the already obtained query $ {\bf{Q}} = \{ {\bf{q}}_{i} \}$ is enhanced by decomposing each query into the K-subspace ${\bf{\Phi}}$ and the complementary subspace using each basis vector ${\bf{\phi}}_{p}$ of the K-subspace.
\begin{eqnarray}
\label{eq:1}
{\bf{q}}_{i} = {\bf{q}}_{i}^{\parallel} + {\bf{q}}_{i}^{\perp},
~~{\bf{q}}_{i}^{\parallel} = {\bf{\Phi}} {\bf{\Phi}}^{T} {\bf{q}}_{i},
~~{\bf{q}}_{i}^{\perp} = ( {\bf{I}} - {\bf{\Phi}} {\bf{\Phi}}^{T}) {\bf{q}}_{i},
\end{eqnarray}
here, ${\bf{q}}_{i}^{\parallel}$ and ${\bf{q}}_{i}^{\perp}$ are the decomposed queries based on the K-subspace ${\bf{\Phi}}$ and its complementary subspace.

Our Q-Perturbation generates a perturbed query ${\hat{\bf{q}}}_{i}$ that enhances the components belonging to the K-subspace, i.e., enhances object information retained in the original query.
This enhancement by the Q-Perturbation is given by
\begin{eqnarray}
\label{eq:4}
{\hat{\bf{q}}}_{i} = {\bf{q}}_{i} + \alpha {\bf{q}}_{i}^{\parallel},
\end{eqnarray}
where $\alpha$ is the parameter that control the perturbation magnitude.

This enhancement could be seen in that relevant queries to an object are automatically selected and then enhanced to have more object information, as the decomposition with the K-subspace has a mechanism to extract only object information retained in the original queries.
Thus, Q-Perturbation enhances object awareness of V\&L models without destroying the weights and structures of the V\&L models, resulting in high inheritability.

\subsection{Extension to Multiple Objects}
\label{sec:multiple_objs}
So far, we have discussed Q-Perturbation on a single object. 
Our proposed Q-Perturbation can be easily extended to the case of multiple objects, i.e., it has high flexibility.
For Q-Perturbation with multiple objects, the proposed method generates the object-aware K-subspace for each object, and then decomposition and enhancement for each query are performed based on these obtained K-subspaces.

Let ${\bf{\Phi}}_{b}$, ${\bf{q}}_{i,b}^{\parallel}$, and ${\bf{q}}_{i,b}^{\perp}$ be the K-subspace corresponding to $b$-th object, query components belonging to the $b$-th K-subspace, and query components orthogonal to the $b$-th K-subspace, respectively.
Here, $b$ is a subscript to distinguish each detected object in an image.
Formally, Q-Perturbation for multiple K-subspaces can be expressed as follows.

\begin{eqnarray}
\label{eq:5}
{\hat{\bf{q}}}_{i} = {\bf{q}}_{i} + \alpha \sum_{b} w(S_b) {\bf{q}}_{i,b}^{\parallel} 
= {\bf{q}}_{i} + \alpha \sum_{b} w(S_b) {\bf{\Phi}}_{b} {\bf{\Phi}}_{b}^{T} {\bf{q}}_{i},
\end{eqnarray}
where $S_b$ and $w(S_b)$ are the area of the detected bounding box and the weight function for $b$-th detected object. 

In this paper, for simplicity, the weight function is given by $w({\bar{S}}_{b})=\beta+\gamma \bar{S}_b$, where $\beta$, $\gamma$ and $\bar{S}$ are the adjustment parameter and normalized area, respectively.
The normalized area $\bar{S}_b=S_b/S_{I}$ is calculated by dividing the whole area of each bounding box by the area $S_{I}$ of the corresponding image.
Note that, in this paper, no exhaustive search was performed for $\beta$ and $\gamma$, and either $ \{ 0,\pm1,\pm0.5 \} $ was used.

\subsection{Beyond the Q-Perturbation Module for Q-Former}
In previous sections, our Q-Perturbation has been described for the case of Q-Former-based model~\cite{BLIP2BootstrappingLanguageImagea}.
Finally, we discuss its extension to other V\&L models and its potential for tasks other than cross-modal image-text retrieval.

\subsubsection{Extension to other pre-trained V\&L models.}
\begin{figure}[t]
    \centering
    \begin{minipage}[c]{.4\hsize}
        \hspace{4pt}
        \includegraphics[width=0.9\columnwidth]{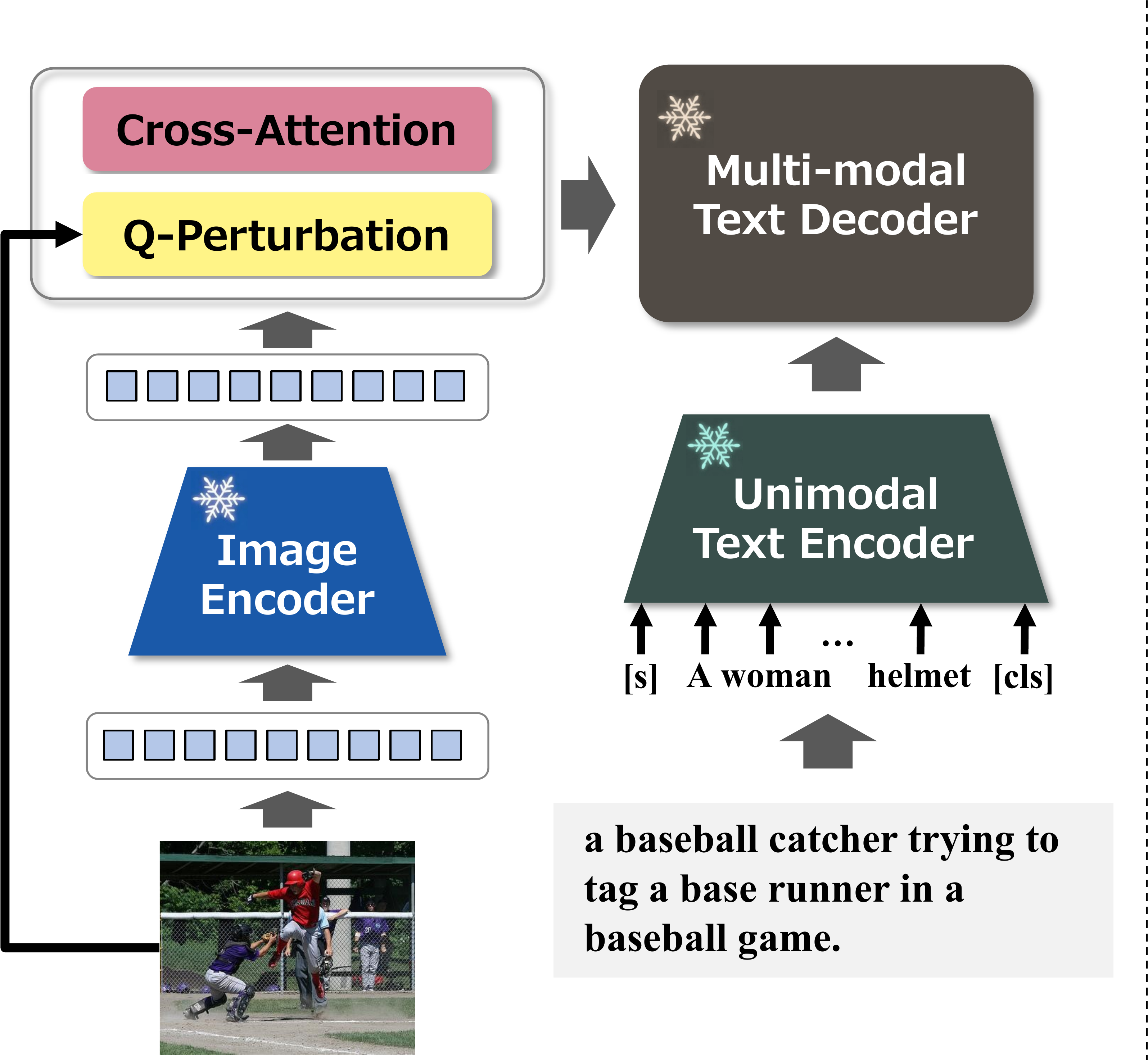}
        \subcaption{COCA with the Q-Perturbation.}
    \end{minipage}
    \begin{minipage}[c]{.4\hsize}
        \hspace{2pt}
        \includegraphics[width=0.9\columnwidth]{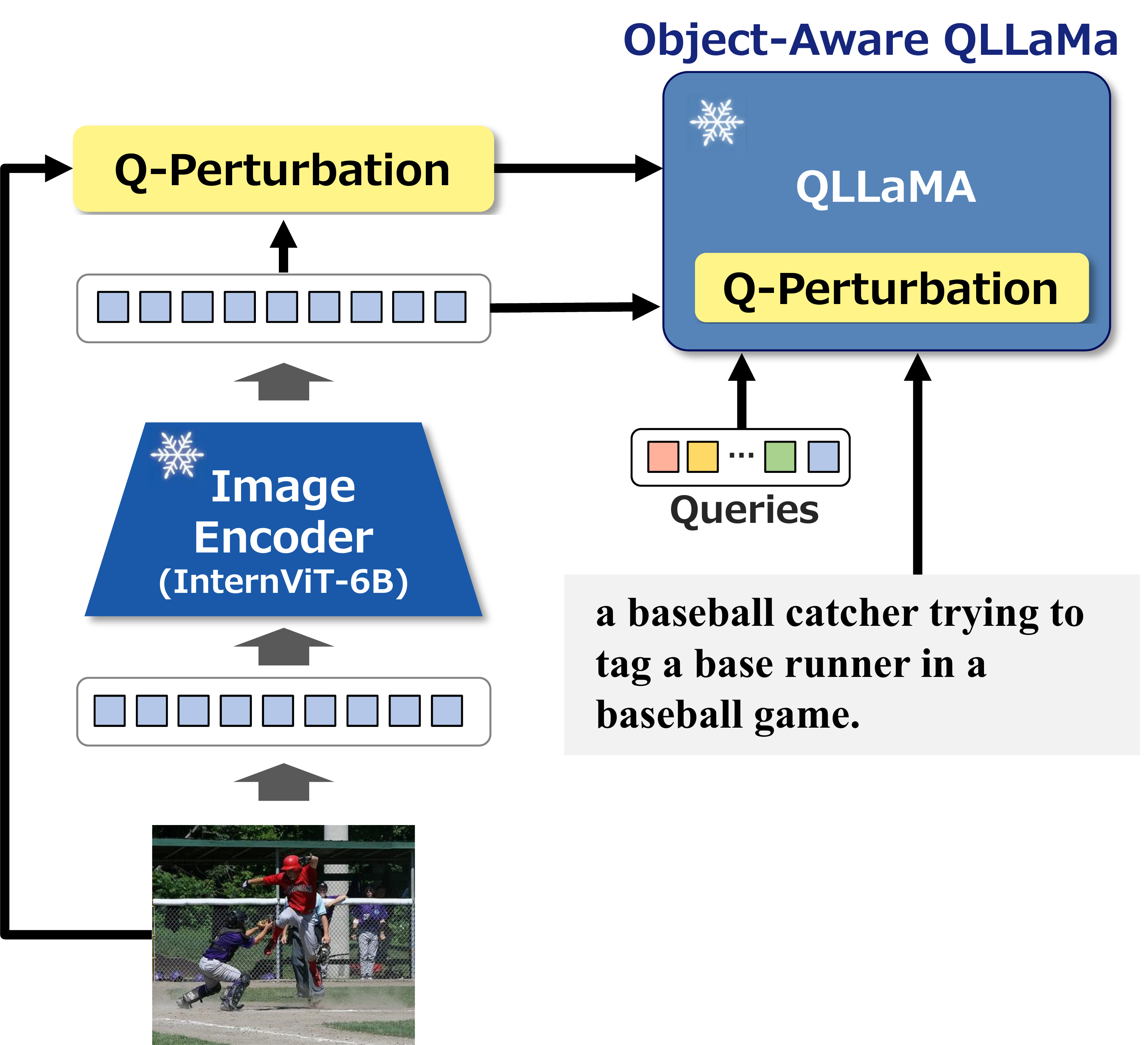}
        \subcaption{InternVL with the Q-Perturbation.}
    \end{minipage}
    \caption{Extension to other pre-trained V\&L models. Our method is applicable to modern V\&L models such as COCA~\cite{yu2022coca} and InternVL~\cite{chen2023internvl}.} 
    \label{fig:variousVLM}
\end{figure}

Many pre-trained V\&L models for cross-modal image-text retrieval have been proposed. It is expected that many more pre-trained models will be proposed in the future.
Our proposed Q-Perturbation module is a general and versatile approach to perturb query in cross-attention using the localization cues from object detection and the obtained key features. 
In this sense, the proposed Q-Perturbation is applicable to other existing V\&L models such as COCA~\cite{yu2022coca} and InternVL~\cite{chen2023internvl}, as shown in Fig.~\ref{fig:variousVLM}. 
As discussed later, the proposed method with the existing V\&L models allows for cross-modal image-text retrieval that is aware of smaller objects.

\subsubsection{Other Tasks with Our Q-Perturbation.}
\label{sec:other_tasks}
In general, the Q-Former proposed in BLIP2 is used for other tasks, such as image captioning by using LLMs in the latter additional stage. 
In this sense, our proposed Q-Perturbation can be used to re-present the existing captions more object-aware. 
This extension suggests new possibilities for using pre-trained V\&L models based on human perception.

\section{Experiments}
\label{sec:exp}
\subsection{Settings}
\subsubsection{Datasets and Experimental Protocols.}
We use two widely used benchmark datasets, i.e., Flickr-30K~\cite{young2014image} and MSCOCO~\cite{chen2015microsoft, lin2014microsoft}, and fine-grained extensions of the two datasets, i.e., Flickr-FG and COCO-FG~\cite{RethinkingBenchmarksCrossmodal}.
We adapt the commonly used Karapathy split~\cite{karpathy2015deep} for all the datasets.

{\it The Flickr 30K} dataset has 1,014 validation images and 1,000 test images.
{\it The MSCOCO} dataset has 5,000 images for validation and test, respectively.
Each image has five description texts. Therefore, there are 5,000 (=1,000 images $\times$ 5 texts) and 25,000 (=5,000 images $\times$ 5 texts) test image-text pairs for Flickr-30K and MSCOCO, respectively.

{\it Flickr-FG and COCO-FG} are extensions of the above two datasets.
Flickr-FG and COCO-FG replaced original description texts with fine-grained descriptions.
These datasets have five fine-grained texts for an image, as with Flickr-30K and MSCOCO.
Therefore, there are 5,000 and 25,000 test image-text pairs, as well.

We conducted text-to-image (T2I) and image-to-text (I2T) tasks; the T2I task is to find the paired image of an input text, and the I2T task is vice versa.

\subsubsection{Evaluation Metrics.}
We use the standard evaluation metric, Recall@K (R@K).
R@K is a ratio of correct retrievals to all retrievals.
Here, the correct retrieval is identified by whether the paired image or texts are in the top K retrieval results.
Following the previous studies, K is set to 1, 5, 10.

We also use mean Recall@K (mR@K), which considers the size of objects in each image.
mR@K is a harmonic mean of multiple R@Ks. 
As outlined in Fig.~\ref{fig:analysis} and elaborated on later, the retrieval difficulty depends on the object size in the image. However, traditional R@K does not consider object size and cannot correctly assess this challenge. To alleviate this problem, we propose to use an object size-aware evaluation metric, mean R@K, which is a harmonic mean of R@Ks on subsets split by object size.
Here, each R@K is calculated on a subset of all text-image pairs, where each subset is determined by the largest normalized area $\bar{S}$ (please see Sec.~\ref{sec:multiple_objs}) of detected objects in each image.
In this paper, we generate ten subsets by splitting largest areas by every 10\% and calculate a harmonic mean (mR@K) of the ten R@Ks calculated on the subsets.

\subsubsection{Implementation Details.}
We use bounding boxes given by Flickr-Entities~\cite{flickrentitiesijcv} and COCO-Entities~\cite{cornia2019show}.
Flickr-Entities and COCO-Entities have bounding boxes corresponding to each text.
Flickr-Entities generates boxes manually by annotators, and COCO-Entities generates boxes semi-automatically, i.e., boxes are detected by Faster R-CNN~\cite{ren2015faster} and matched nouns in each text by manually defined rules.
We also use boxes that are detected automatically by CO-DINO~\cite{zong2023detrs}. 
Note that, object detection and image feature extraction can be carried out in advance for the T2I task, as they do not depend on a retrieval input text.

To split datasets and calculate mR@K, we use Flickr-Entities's bounding boxes for Flickr-30K and Flickr-FG.
For COCO and COCO-FG, we use bounding boxes detected by CO-DINO instead of COCO-Entities, as COCO-Entities are built by the traditional detector, Faster R-CNN.

We applied the proposed Q-Perturbation to all cross-attentions in the Q-Former of BLIP2.
We used Eva-CLIP~\cite{sun2023eva} base BLIP2 finetuned on COCO validation data~\cite{li-etal-2023-lavis, BLIP2BootstrappingLanguageImagea}.
Following the previous study~\cite{BLIP2BootstrappingLanguageImagea}, we apply the re-ranking technique; we first select 64 candidates by image-text contrastive (ITC) and then re-ranked them by image-text matching (ITM).

We also evaluated our method combined with COCA and InternVL-G models~\cite{yu2022coca,chen2023internvl}.
    We used the ViT-L/14~\cite{vit} based COCA model published by the OpenCLIP repository~\cite{openclip}, and the InternVL-14B-224px model~\cite{chen2023internvl}.
Our Q-Perturbation is applied to the last cross-attention for COCA and to all cross-attentions in the QLLaMA for InternVL-G (see the supplementary material A for more detail).

Q-Perturbation has three hyperparameters: 1) perturbation intensity $\alpha$, 2) weight function $w(S_b)$, and 3) dimension of object-aware subspaces.
These parameters were tuned by the grid-search algorithm using validation data with the mR@1.
The intensity $\alpha$ was selected from 2, 4, 6, 8 and 10 for BLIP2 and 0.2, 0.4, 0.6, 0.8 and 1 for COCA and InternVL.
The weight function was selected from five functions: constant value (=1), $\bar{S}_b$, $\bar{S}_b-0.5$, $1-\bar{S}_b$, $0.5-\bar{S}_b$. 
The dimensions of subspaces were determined by using the contribution ratio.
Thus, we tuned the threshold of the contribution ratio from 0.85, 0.9, 0.95, 0.99.

\subsection{Comparative results}
\input{tables/eccv_tab1}

\begin{figure}[t]
    \centering
    \centering    
    \begin{minipage}[t]{.495\columnwidth}
        \includegraphics[width=\columnwidth]{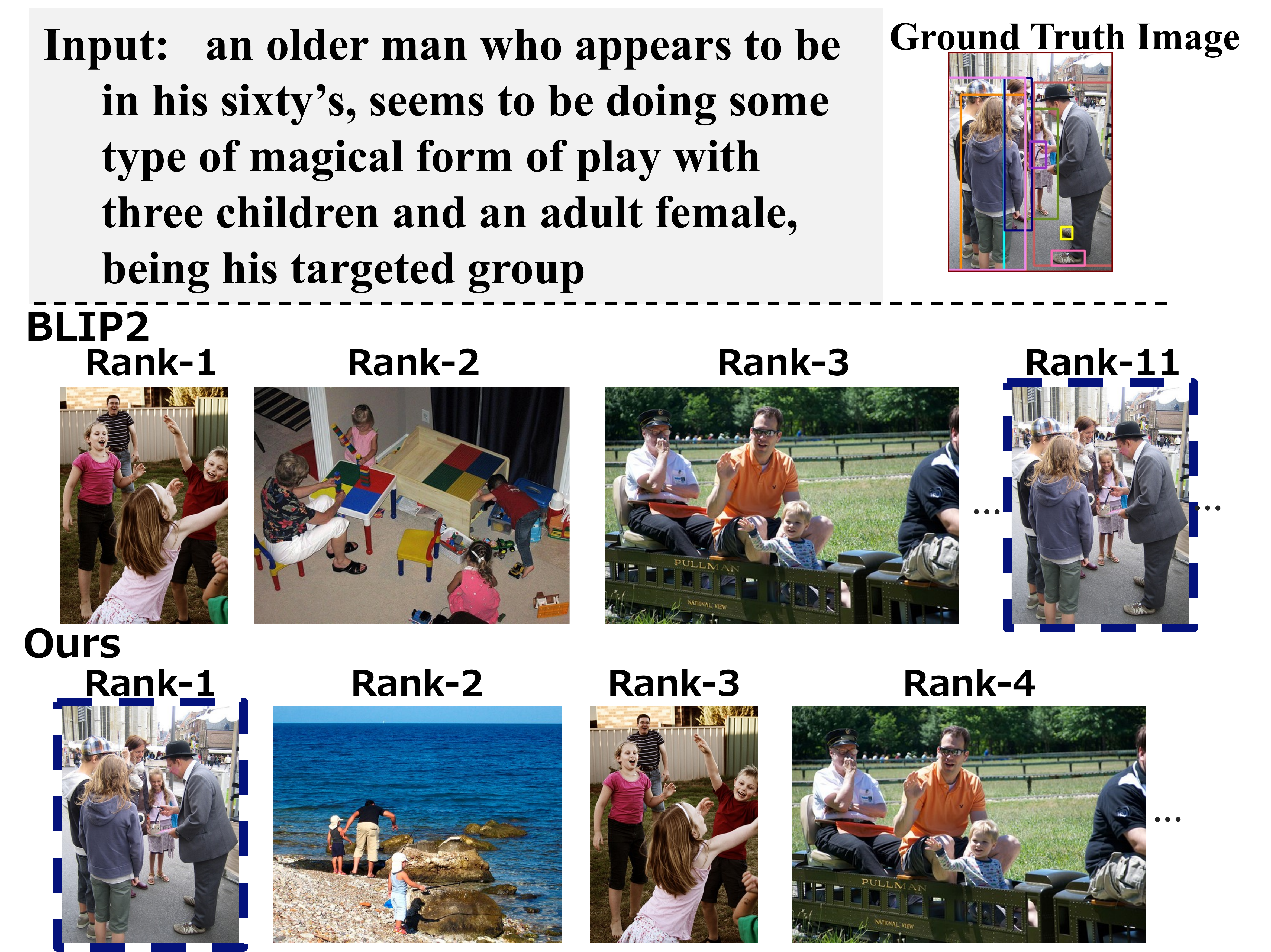}
        \subcaption{}
    \end{minipage}
    \begin{minipage}[t]{.495\columnwidth}
        \includegraphics[width=\columnwidth]{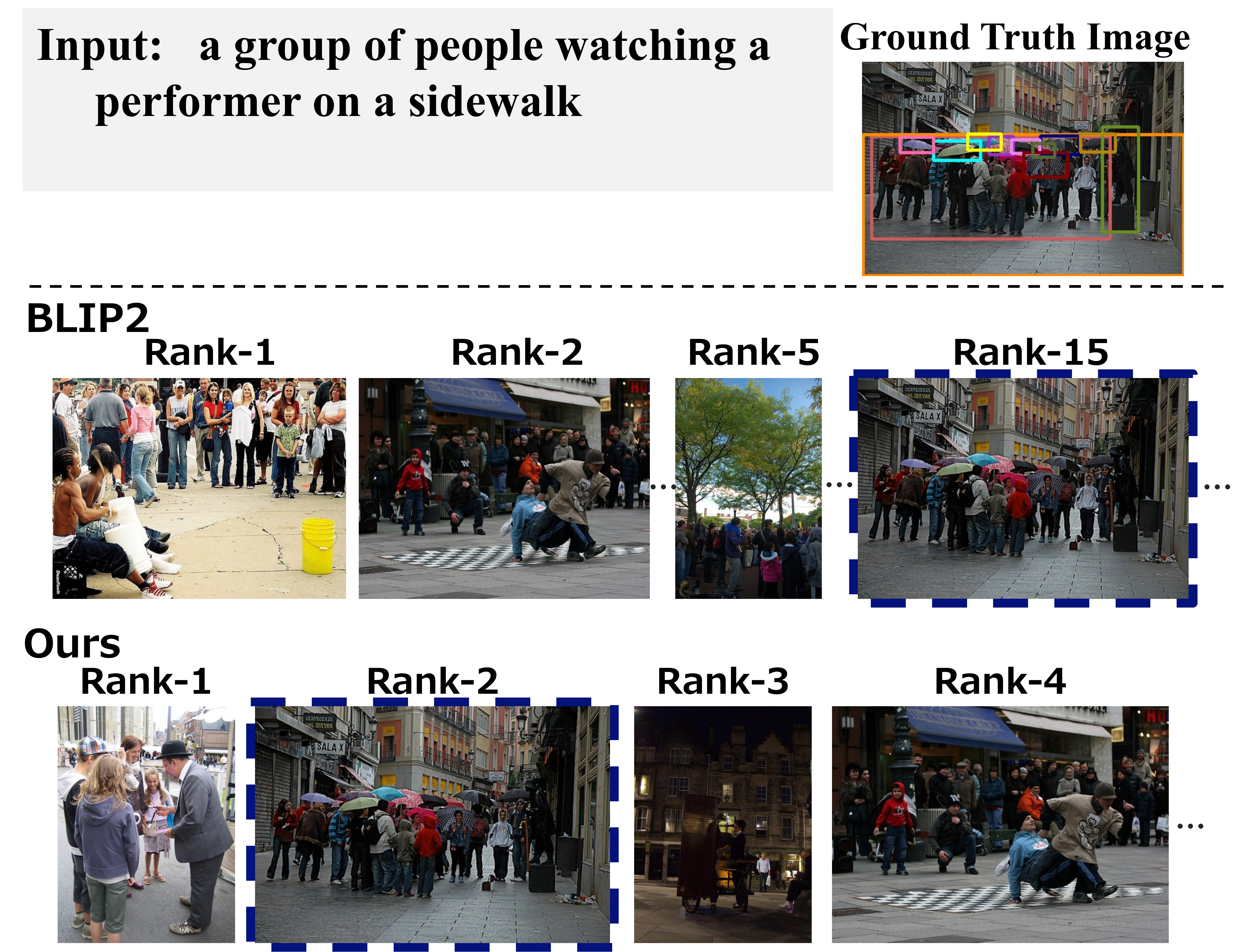}
        \subcaption{}
    \end{minipage}
    \caption{Examples of image retrieval results by BLIP2~\cite{BLIP2BootstrappingLanguageImagea} and our method.} 
    \label{fig:results_example}
\end{figure}

\subsubsection{Results for small objects.}
Table~\ref{tab:result_tab1} shows the evaluation results for small objects and the overall of each dataset.
The proposed Q-Perturbation improves the retrieval performance of small objects, which in turn improves the overall performance.
The conventional method would have had difficulty in considering small objects in the image.
The results suggest that our method mitigates this difficulty by enhancing object awareness of the conventional methods.

Figure~\ref{fig:results_example} shows examples of image retrieval results by BLIP2-ITC and BLIP2-ITC with Q-Perturbation.
We can see that our method selects the correct image at a higher rank than the original BLIP2. 
This is mainly due to the advantage of our object-aware mechanism; the proposed method efficiently utilizes information from small objects, such as ``an older man'', and ``a performer'' for Fig.~\ref{fig:results_example} (a) (b), respectively.  
\input{tables/eccv_tab2}

Table~\ref{tab:results_tab2} shows comparative results with additional evaluation metrics.
We can confirm the effectiveness of our method again, as our method shows competitive results.
Furthermore, our method stably improves performance even with noisy bounding boxes obtained by a detector automatically.
This property is helpful in applying our method to real-world applications.

\begin{figure}[t]
    \centering    
    \begin{minipage}[c]{.49\hsize}
        \centering
        \includegraphics[width=\linewidth]{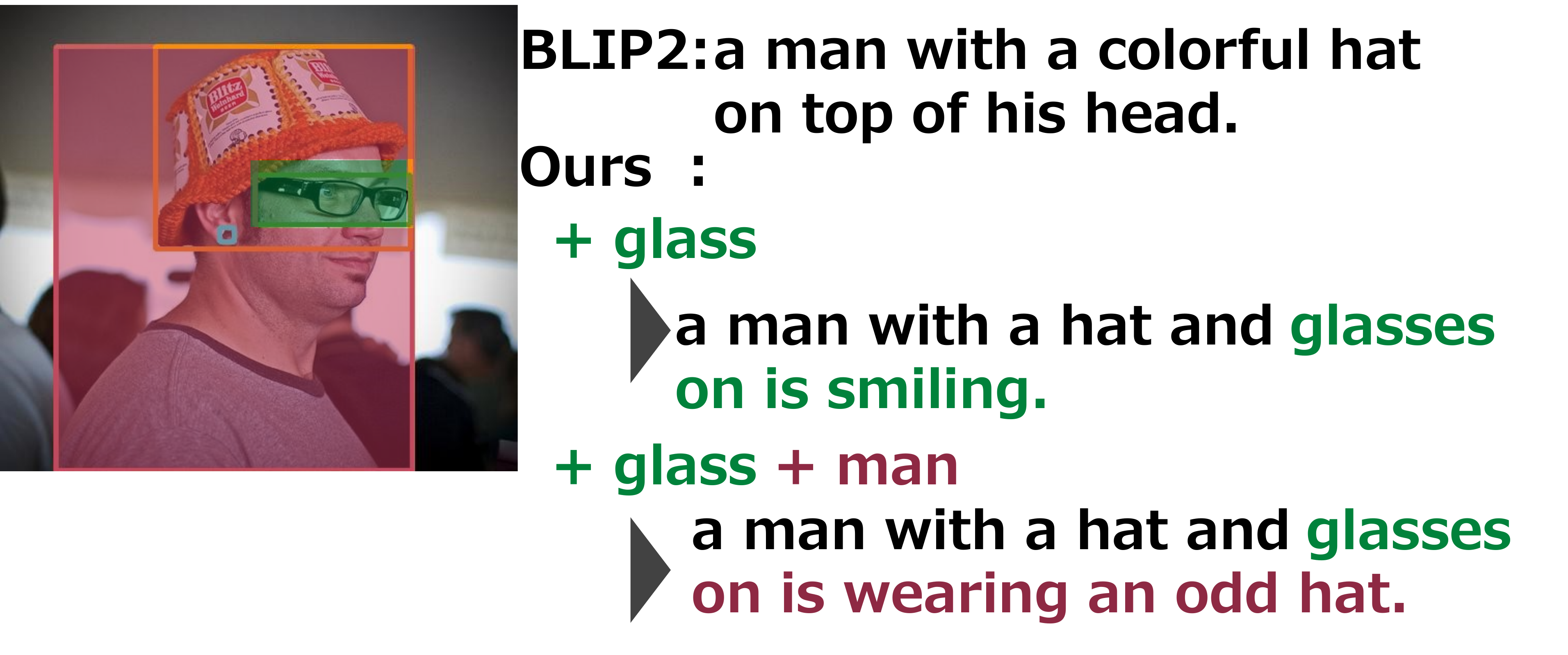} 
    \end{minipage} 
    \begin{minipage}[c]{.49\hsize}
        \centering
        \includegraphics[width=\linewidth]{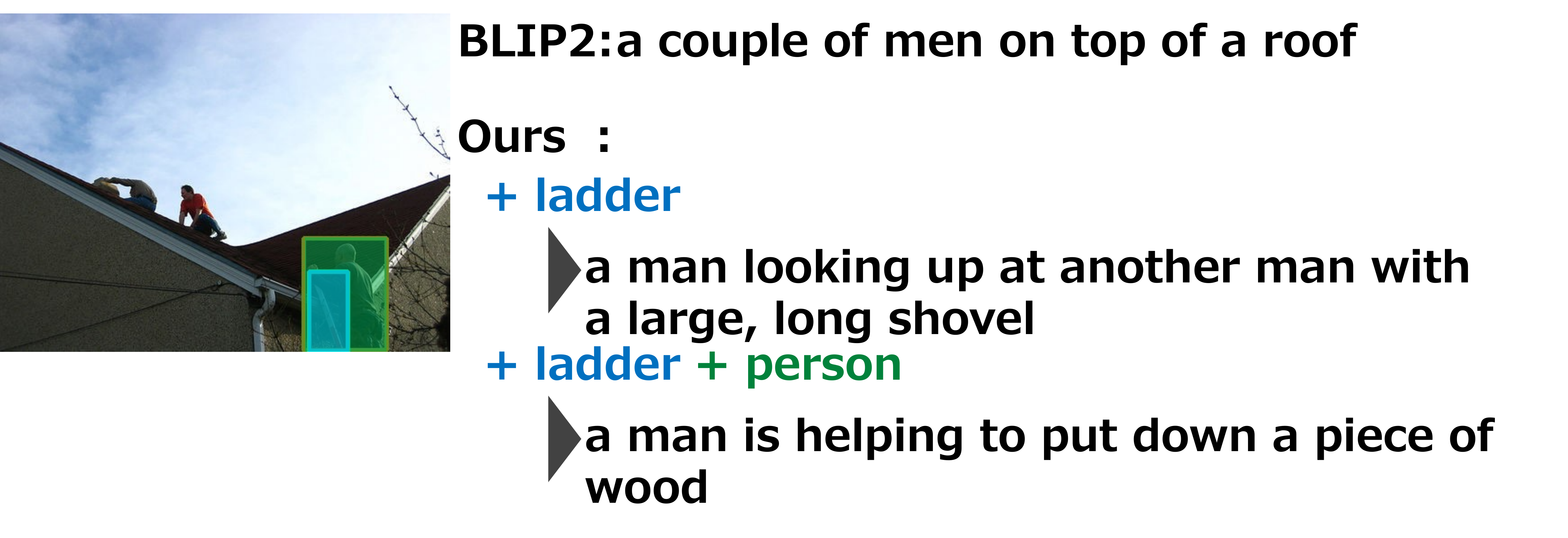}
    \end{minipage}
    \caption{Image captioning results. ``$+$ word'' means enhancing information of the region corresponding to the word.}
    \label{fig:chat_example}
\end{figure}

As we discussed in Sec.~\ref{sec:other_tasks}, our Q-Perturbation is applicable to other tasks, such as image captioning, as our method is plugged into a pre-trained V\&L model.
We carried out image captioning to visualize the effect of Q-Perturbation, as shown in Fig.~\ref{fig:chat_example}.
It can be seen that output captions become object-aware compared with the original BLIP2s' results by emphasizing the corresponding objects, such as ``glass'', with our Q-Perturbation.
This object-aware property of our method helps in improving the retrieval performance of a V\&L model.

\subsubsection{Overall results.}
We then discuss performance comparisons between our method and various cross-modal image-text retrieval methods.

{\noindent - \bf Baselines:} We compared our method with 1) object-aware models; SCAN~\cite{lee2018stacked}, IMRAM~\cite{IMRAMIterativeMatching}, SHAN~\cite{StepWiseHierarchicalAlignment}, NAAF~\cite{zhang2022negative}, and 2) V\&L pre-trained models; CLIP~\cite{clip}, ALBEF~\cite{albef}, UNITER~\cite{chen2020uniter}, BEIT-3~\cite{beit}, COCA~\cite{yu2022coca}, InternVL~\cite{chen2023internvl}, BLIP2~\cite{BLIP2BootstrappingLanguageImagea}.

{\noindent - \bf Results:}
Table~\ref{tab:results_tab3} shows comparative results with various conventional methods.
Our method archives competitive results compared with state-of-the-art methods.
These results suggest that our Q-Perturbation enhances object awareness of the pre-trained V\&L model while inheriting its impressive performance (for more comparison with simple baselines, see supplementary material B).

\subsection{Results with Other State-of-The-Art V\&L Models}
Our proposed Q-Perturbation can be plugged into any model, including cross-attention layers.
To confirm the versatility of our method, we apply Q-Perturbation to two V\&L models, COCA and InternVL.
Table~\ref{tab:results_tab4} shows comparative results.
We can see that Q-Perturbation is highly versatile as our method improves retrieval performance.
COCA w/Q-pert. has a slight improvement. This is because the impact of the proposed method is small, as a cross-attention layer is only placed at the end of the vision encoder.
Input features to QLLaMA, which is the cross-modal projector of InternVL, or cross-attentions may focus on the global context, as QLLaMA is followed by the large vision encoder (6B parameters ViT model).
Even in such a difficult situation, the proposed method could enhance object-awareness and improve performance.

\subsection{Sensitivity on hyperparameters}
\input{tables/eccv_tab3}

\input{tables/eccv_tab4}
\input{tables/eccv_tab5}
We analyze the sensitivity of the performance to the hyperparameters, including the weight function, scale factor, and dimension of subspaces, i.e., threshold of contribution ratio for PCA.
In this experiment, we use bounding boxes by Flickr-Entities.
Table~\ref{tab:scale_sensitivity} shows the evaluation results with varying the hyperparameters.
The proposed method has a high performance by adequately selecting the parameters, although the proposed method has low sensitivities to the hyperparameters. 
In this paper, we selected the hyperparameters from the manually set values using validation data. 
It would be an excellent direction to learn the hyperparameters in the future if learning data were available. 

\section{Limitations}
The key idea behind our method is to enhance object information in a V\&L model's cross-attention features following an image encoder.
It would, therefore, be problematic if object information had disappeared entirely in the image encoder.

This paper uses all bounding boxes in each image.
It is an excellent future direction to filter bounding boxes or adjust the scale $\alpha$ according to an input text to improve retrieval performance further. 
Note that this approach increases computational cost at the retrieval stage, as we need to extract bounding boxes and process a neural network, including cross-attentions, after receiving the input.
This direction would be a trade-off between scalability and retrieval performance.

\section{Conclusions}
In this paper, we proposed an object-aware query perturbation for cross-modal image-text retrieval.
The key is to use query perturbation to focus on small target objects by enhancing the query weights with keys corresponding to object regions in the cross-attention module.
The proposed method is applicable to various V\&L models based on cross-modal projection, including COCA, BLIP2, and InternVL.
Comprehensive experiments on four public datasets demonstrate the effectiveness of the proposed method.

\bibliographystyle{splncs04}
\bibliography{main}

\newpage

\titlerunning{Supplementary Material for Object-Aware Query Perturbation}
% A,B,C,...
\renewcommand{\thetable}{\Alph{table}} 
\renewcommand\thesection{\Alph{section}}
\renewcommand\thefigure{\Alph{figure}}

\setcounter{section}{0}

\include{supp}

\end{document}

%% file: tables/eccv_tab1.tex
\begin{table}[t]
    \centering
    \caption{Result for the T2I retrieval task. Q-Pert. (E) and (D) are the results of the proposed method using bounding boxes produced by Flickr-/COCO-Entities (E), and detected by CO-DETR (D), respectively. \textasciitilde10\% means R@1 calculated for data with only tiny objects, i.e., the object area is smaller than 10\% to the corresponding image area. The best and second-best are highlighted in bold and underlined, respectively.}
    \label{tab:result_tab1}
    \centering
    \scalebox{0.72}{
        \begin{tabular}{lrrrrrrrrrrrr} 
            \toprule
              &  \multicolumn{3}{c}{Flickr-30K}& \multicolumn{3}{c}{Flickr-FG} & \multicolumn{3}{c}{COCO-5K}& \multicolumn{3}{c}{COCO-FG}\\
            \midrule
                & \textasciitilde 10\%&R@1& mR@1 & \textasciitilde 10\%& R@1&mR@1  & \textasciitilde 10\%& R@1& mR@1 & \textasciitilde 10\%& R@1&mR@1  \\ 
            \cmidrule(lr){2-4} \cmidrule(lr){5-7}   \cmidrule(lr){8-10} \cmidrule(lr){11-13}  
              BLIP2        
      & \second{81.33}& 89.76 & 88.75
& \second{85.33}& 92.50 &91.79
 & 59.87
& 68.25
& 71.09
& 63.93
& 72.68
&75.71
\\ 
              ~~w/Q-Pert.(E)   & \first{84.00}& \second{89.82}& \second{89.10}& \first{86.67}& \second{92.56}&\second{91.98}& \first{60.17}& \second{68.34}& \second{71.14}& \second{64.00}& \second{72.75}&\second{75.77}\\
              ~~w/Q-Pert.(D)   & \first{84.00}& \first{89.86}& \first{89.11}& \first{86.67}& \first{92.66}&\first{92.06}& \second{60.09}& \first{68.35}& \first{71.15}& \first{64.06}& \first{72.78}&\first{75.78}\\ \bottomrule
        \end{tabular}
    }
\end{table}

%% file: tables/eccv_tab2.tex
\begin{table}[t]
    \caption{Results on four datasets for T2I and I2T retrieval tasks. Q-Pert. (E) and (D) are the results using bounding boxes produced by Flickr-/COCO-Entities (E), and detected by CO-DINO (D), respectively. The best and second-best are highlighted in bold and underlined, respectively.}
    \label{tab:results_tab2}
    \centering
    \scalebox{0.72}{
        \begin{tabular}{lrrrrrrrrrrrr}
        \toprule
               & \multicolumn{6}{c}{T2I} & \multicolumn{6}{c}{I2T}\\
         \cmidrule(lr){2-7} \cmidrule(lr){8-13}
             &  R@1&  R@5&  R@10&  mR@1&  mR@5& mR@10 & R@1& R@5& R@10& mR@1& mR@5&mR@10 \\
         \cmidrule(lr){2-4} \cmidrule(lr){5-7} \cmidrule(lr){8-10} \cmidrule(lr){11-13}
         & \multicolumn{12}{c}{Flickr-30K}\\
        \midrule
               BLIP2         & 89.76 & 98.18 & 98.96 & 88.75 & 97.85 & \second{98.66}& 97.60 & 100.00& 100.00& 97.23 & 100.00 & 100.00 
\\
               ~w/Q-Pert. (E)& \second{89.82}& \first{98.20}& \second{99.04}& \second{89.10}& \first{97.89}& 98.60 
& \first{98.00}& 100.00& 100.00& \first{97.55}& 100.00& 100.00
\\
               ~w/Q-Pert. (D)&  \first{89.86}&  \first{98.20}&  \first{99.06}&  \first{89.11}&  \second{97.86}&  \first{98.76}& \second{97.70}& 100.00& 100.00& \second{97.29}& 100.00&100.00
\\
        \midrule
          & \multicolumn{12}{c}{COCO 5K}\\
        \midrule
               BLIP2         & 68.25
& \second{87.74}& \first{92.67}& 71.09
& \second{89.82}& \first{94.12}& 85.32& \first{96.94}& \first{98.38}& 87.43
& \first{97.78}& \second{98.86}\\
               ~w/Q-Pert. (E)& \second{68.34}& \first{87.76}& 92.63 & \second{71.14}& \first{89.84}& \second{94.11}& \first{85.52}& \second{96.90}& \first{98.38}& \second{87.52}& \second{97.74}& \first{98.87}\\
               ~w/Q-Pert. (D)&  \first{68.35}&  87.72&  \second{92.65}&  \first{71.15}&  89.79&  \second{94.11}& \second{85.44}& 96.88
& 
\second{98.36}& \first{87.69}& 97.71&98.85
\\
        \midrule
          & \multicolumn{12}{c}{Flickr-FG}\\
        \midrule
               BLIP2         & 92.50 & \second{99.00}& 99.46 & 91.79 & 98.83 & 99.45 
& \second{98.50}& 100.00& 100.00& 98.74& 100.00& 100.00
\\
               ~w/Q-Pert. (E)& \second{92.56}& \second{99.00}& \first{99.56}& \second{91.98}& \first{98.95}& \first{99.53}& \first{98.6}& 100.00& 100.00& \first{98.87}& 100.00& 100.00
\\
               ~w/Q-Pert. (D)&  \first{92.66}&  \first{99.06}&  \second{99.48}&  \first{92.06}&  \second{98.88}&  \second{99.47}& \first{98.6}& 100.00& 100.00& \second{98.85}& 100.00&100.00
\\
        \midrule
          & \multicolumn{12}{c}{COCO-FG}\\
        \midrule
              BLIP2         & 72.68
& \second{90.22}& \second{94.24}& 75.71
& 92.17
& \second{95.52}& 87.58
& 97.64
& \second{98.94}& 89.83
& 98.42
& \second{99.32}\\
              ~w/Q-Pert. (E)& \second{72.75}& \first{90.23}& 94.22
& \second{75.77}& 92.17
& 95.49
& \first{87.82}& \second{97.64}& 98.92
& \first{89.98}& \second{98.44}& \first{99.37}\\
              ~w/Q-Pert. (D)& \first{72.78}& \second{90.22}& \first{94.30}& \first{75.78}& 92.17& \first{95.58}& \second{87.66}& \first{97.70}& \first{98.98}& \second{89.87}& \first{98.45}&\first{99.37}\\ \bottomrule
        \end{tabular}
    }
\end{table}

%% file: tables/eccv_tab3.tex
\begin{table}[t]
    \caption{Comparative results with baselines. The best and second-best are highlighted in bold and underlined, respectively.}
    \label{tab:results_tab3}
            \centering
    \scalebox{0.72}{
        \centering
        \begin{tabular}{lrrrrrrrrrrrr}
        \toprule
             &  \multicolumn{6}{c}{Flickr-30K}&  \multicolumn{6}{c}{COCO 5K}\\
        \cmidrule(lr){2-7} \cmidrule(lr){8-13}
             &  \multicolumn{3}{c}{T2I
}&  \multicolumn{3}{c}{I2T
}&  \multicolumn{3}{c}{T2I
}& \multicolumn{3}{c}{I2T
}\\
         \cmidrule(lr){2-4} \cmidrule(lr){5-7} \cmidrule(lr){8-10} \cmidrule(lr){11-13}        
             & R@1& R@5& R@10
& R@1& R@5& R@10
& R@1& R@5& R@10
& R@1& R@5&R@10
\\
        \midrule
             SCAN~\cite{lee2018stacked}&  48.60 & 77.70 & 85.20 
& 67.40 & 90.30 & 95.80 
& 38.60 & 69.30 & 80.40 
& 50.40 & 82.20 & 90.00 
\\
             IMRAM~\cite{IMRAMIterativeMatching}&  53.90 & 79.40 & 87.20 
& 74.10 & 93.00 & 96.60 
& 39.70 & 69.10 & 79.80 
& 53.70 & 83.20 & 91.00 
\\
             SHAN~\cite{StepWiseHierarchicalAlignment}& 55.30 & 81.30 & 88.40 
& 74.60 & 93.50 & 96.90 
& - &	- &	-
&	- &	- &- 
\\
             VSE~\cite{VSEImprovingVisualSemantic}& 76.10 & 94.50 & 97.10 & 88.70 & 98.90 & \second{99.80}& 52.70 & 80.20 & - &	68.10 & 90.20 & - \\
             NAAF~\cite{zhang2022negative}&  61.00 & 85.30 & 90.60 
& 81.90 & 96.10 & 98.30 
& 42.50 & 70.90 & 81.40 
& 58.90 & 85.20 & 92.00 
\\
             CLIP~\cite{clip}& 68.70 & 90.60 & 95.20 
& 88.00 & 98.70 & 99.40 
& - &	- &	-
&	- &	- &	- 
\\
             ALBEF~\cite{albef}& 82.80 & 96.30 & 98.10 
& 94.10 & 99.50 & 99.70 
& 60.70 & 84.30 & 90.50 
& 77.60 & 94.30 & 97.20 
\\
             UNITER~\cite{chen2020uniter}&  68.70 & 89.20 & 93.90 
& 83.60 & 95.70 & 97.70 
& 52.90 & 79.90 & 88.00 
& 65.70 & 88.60 & 93.80 
\\
             BEIT-3~\cite{beit}& 81.50 & 95.60 & 97.80 
& 94.90 & \second{99.90}& \first{100.00} 
& 67.20 & 87.70 & \first{92.80}& 84.80 & 96.50 & 98.30 
\\
             COCA~\cite{yu2022coca}& 77.64 & 94.00 & 96.48 & 90.90 & 98.20 & 99.20  & - &	- &	-  &	- &	- &	- \\
             % COCA w/Q-Pert.& & & & & & & & & & & &\\
             InternVL-G~\cite{chen2023internvl}& 84.78 & 97.02& 98.38& 94.90& 99.90& \first{100.00}& - &	- &	- &	- &	- &	- \\
             % InternVL-G w/Q-Pert.& & & & & & & & & & & &  \\
        \midrule             
             BLIP2~\cite{BLIP2BootstrappingLanguageImagea}& 89.76 & \second{98.18}& 98.96 
& 97.60 & \first{100.00}& \first{100.00} 
& 68.25
& \second{87.74}& \second{92.67}& 85.32 & \first{96.94}& \first{98.38}\\
             ~~w/Q-Pert.(E)& \second{89.82}& \first{98.20}& \second{99.04}& \first{98.00}& \first{100.00} & \first{100.00} 
& \second{68.34}& \first{87.76}& 92.63 
& \first{85.52}& \second{96.90}& \first{98.38}\\
 ~~w/Q-Pert.(D)& \first{89.86}& \first{98.20}& \first{99.06}& \second{97.70}& \first{100.00} & \first{100.00} 
& \first{68.35}& 87.72& 92.65
& \second{85.44}& 96.88&
\second{98.36}\\ \bottomrule
        \end{tabular}
    }
    %\caption{Comparative results with baselines.}
    %\label{tab:results_tab3}
\end{table}

%% file: tables/eccv_tab4.tex
\begin{table}[t]
    \caption{Results with the method of Q-perturbation with two V\&Ls. }
    \label{tab:results_tab4}
    \centering
    \scalebox{0.72}{
        \centering
        \begin{tabular}{lrrrrrrrrrrrr}
        \toprule
             &  \multicolumn{6}{c}{Flickr-30K}&  \multicolumn{6}{c}{Flickr-FG}\\
        \cmidrule(lr){2-7} \cmidrule(lr){8-13}
             &  \multicolumn{3}{c}{T2I}&  \multicolumn{3}{c}{I2T}&  \multicolumn{3}{c}{T2I}& \multicolumn{3}{c}{I2T}\\
         \cmidrule(lr){2-4} \cmidrule(lr){5-7} \cmidrule(lr){8-10} \cmidrule(lr){11-13}        
             & R@1& R@5& R@10& R@1& R@5& R@10& R@1& R@5& R@10& R@1& R@5&R@10\\
        \midrule
             COCA& 77.64 & \first{94.00}& 96.48 
& 90.90 & 98.20 & 99.20 
& 82.84 & 96.10 & \first{97.90}& 93.00 & 99.10 &99.70 
\\
             ~~w/Q-Pert.& \first{77.68}& 93.98 & \first{96.50}& \first{91.10}& 98.20& 99.20& \first{82.88}& \first{96.12}& 97.86
& \first{93.20}& 99.10&99.70\\
\hdashline
             InternVL-G& 84.78 & 97.02 & 98.38 
& 94.90 & 99.90 & 100.00& 89.04 & 98.22 & 99.14 
& 97.10 & 99.90 &100.00\\
             ~~w/Q-Pert.& \first{84.82}& 97.02& \first{98.42}& \first{95.40}& 99.90 & 100.00& \first{89.10}& \first{98.26}& 99.14 
& \first{97.20}& 99.90 &  100.00\\
        \midrule
            & mR@1& mR@5& mR@10& mR@1& mR@5& mR@10& mR@1& mR@5& mR@10& mR@1& mR@5&mR@10\\
        \cmidrule(lr){2-4} \cmidrule(lr){5-7} \cmidrule(lr){8-10} \cmidrule(lr){11-13}        
             COCA& 75.86& 92.76& 95.43& 88.89& 96.95& 99.34& 81.78& 94.92& \first{97.47}& 93.08& 98.57&99.77
\\
            ~~w/Q-Pert.& \first{75.91}& \first{92.77}& \first{95.45}& \first{89.03}& 96.95& 99.34& \first{81.83}& \first{94.95}& 97.44& \first{93.16}& 98.57&99.77
\\\hdashline
             InternVL-G& 83.12 & 96.67 & 98.05 
& 94.35 & 99.89 & 100.00& 87.68 & 97.98 & 99.07 
& 97.39 & 99.83 &100.00\\
             ~~w/Q-Pert.& \first{83.15}& 96.67& \first{98.10}& \first{94.75}& 99.89 & 100.00& \first{87.72}& \first{98.02}& 99.07& \first{97.46}& 99.83 &  100.00\\
        \bottomrule
        \end{tabular}
    }
    %\caption{Results with the method of Q-perturbation with two VLMs.}
    %\label{tab:results_tab4}
\end{table}

%% file: tables/eccv_tab5.tex
\begin{table}[t]
        \caption{Sensitivity on hyperparameters, i.e., the weight function, scale factor, and dimension of subspace.}
        \label{tab:scale_sensitivity}
        \centering
    \begin{minipage}[c]{.45\hsize}
        \centering
        \subcaption{Flickr-30K datasets. mR@1.}
        \vspace{-0.3cm}
        \scalebox{0.72}{
        \begin{tabular}{llrrrrr}
        \toprule
            && \multicolumn{5}{c}{scale}\\
        \midrule
            &&  2 &  4&  6&  8& 10\\
        \midrule
            &$\bar{S}_b$ &  \first{89.76}&  89.56&  89.68&  \second{89.72}& 89.66\\
            &1-$\bar{S}_b$ &  \second{89.74}&  89.52&  89.68&  \first{89.78}& \second{89.68}\\
            &$\bar{S}_b$-0.5 &  89.72&  89.56&  \second{89.76}&  89.50 & 89.52\\
            &0.5-$\bar{S}_b$&  89.72&  \first{89.84}&  89.66&  \second{89.72}& \first{89.70}\\
            &constant&  89.66&  \second{89.74}&  \first{89.82}&  \second{89.72}& 89.44
\\
        \bottomrule
        \end{tabular}
        }
        % \subcaption{mean R@1}
    \end{minipage}
    \begin{minipage}[c]{.45\hsize}
        \centering
        \subcaption{Flickr-30K datasets. mR@1.}
        \vspace{-0.3cm}
        \scalebox{0.72}{
        \begin{tabular}{llrrrrr}
        \toprule
            && \multicolumn{5}{c}{Contribution ratio}\\
        \midrule
            &&  0.80&  0.85&  0.90&  0.95& 0.99\\
        \midrule
            &$\bar{S}_b$ &  89.64&  89.68&  \second{89.76}&  89.62& 89.66
\\
            &1-$\bar{S}_b$ &  \first{89.94}&  \first{90.00}&  89.62&  \first{89.82}& \second{89.68}\\
            &$\bar{S}_b$-0.5 &  \second{89.84}&  89.58&  89.54&  89.68& 89.52
\\
            &0.5-$\bar{S}_b$&  89.80&  \second{89.86}&  89.74&  \second{89.74}& \first{89.70}\\
            &constant&  89.56&  89.68&  \first{89.82}&  89.64& 89.44
\\
        \bottomrule
        \end{tabular}
        }
        %\subcaption{mean R@1}
    \end{minipage}
\end{table}

%% file: supp.tex
% ---------------------------------------------------------------
\title{Supplementary Material for \\ ``Object-Aware Query Perturbation \\ for Cross-Modal Image Retrieval''} 

\author{Naoya Sogi \and
Takashi Shibata \and
Makoto Terao
}

\authorrunning{N.~Sogi et al.}

\institute{Visual Intelligence Research Laboratories, NEC Corporation,
Kanagawa, Japan \\
\email{naoya-sogi@nec.com, t.shibata@ieee.org, m-terao@nec.com}\\
}
\maketitle

\section{Experimental details}

\subsection{Evaluation Metrics}
As with the previous works, we used Recall@K, which is a ratio of correct retrievals to all retrievals. 
For the text-to-image (T2I) retrieval task, the correct retrieval is identified by whether the image corresponding to the input text is in the top K retrieval results.
For the image-to-text (I2T) retrieval task, the correct retrieval is identified by whether at least one of the five description texts of the input image is in the top K retrieval results.

Our mean Recall@K is a harmonic mean of multiple Recall@Ks; each of them is calculated with the above procedure on a subset of test data.
We used ten subsets by splitting the largest object areas in each image by every 10\%.

\subsection{Implementation Details}
\subsubsection{Q-Perturbation.}
We applied our Q-Perturbation to the following cross-attention modules: for BLIP2, six cross-attentions in Q-Former; for InternVL, sixteen cross-attentions in QLLaMA and one additional cross-attention after vision encoder; for COCA, one cross attention after vision encoder.
For BLIP2, Q- Perturbation was applied for the image-text contrastive (ITC) phase.
 
For multi-head cross-attention modules, Q-Perturbation is applied to each head.
Algorithm \ref{code1} shows a PyTorch-like pseudo code of Q-Perturbation for each head of cross attentions.
This function is inserted before calculating attention scores. 

\input{supp_tab/psudo_code}

To implement image captioning by BLIP2 with Q-Perturbation, we used the language head following Q-Former~\cite{BLIP2BootstrappingLanguageImagea}.

\subsubsection{Generation of a Key Subspace.}
Let ${\bf K}^{obj}\in \mathbb{R}^{D\times N}$ be a matrix, whose columns are image tokens selected by our Object Key Pooling; each column is a key vector in a cross-attention overlapping a detected object.
Here, $D$ is the dimension of a vector, and $N$ is the number of tokens.

We generate basis vectors $\{\phi_i\}$ of a key subspace by applying Principal Component Analysis (PCA) to $\bf K^{obj}$. Formally, basis vectors are obtained by solving the following eigenvalue decomposition problem:
\begin{align}
    {\bf K^{obj}}{\bf K^{obj}}^T = {\bf\Phi} {\bf\Sigma} {\bf\Phi}^T,
\end{align}
where, diagonal elements ${\sigma_{i,i}}$ of $\bf \Sigma$ are eigenvalues ordering descending order, i.e., $\sigma_{1,1}>\sigma_{2,2,}>...>\sigma_{min(D,N),min(D,N)}$, and eigenvectors ${\bf \Phi}=[\phi_1, \phi_2, ..., \phi_{min(D,N)}]$ are candidates for basis vectors of a key subspace.
We select the first $p$ eigenvectors, $[\phi_1, \phi_2, ..., \phi_{p}]$, as basis vectors.
The number of basis vectors $p$ is set based on the following contribution ratio:
\begin{align}
    c_p = \frac{\sum_{i=1}^p{\sigma_{i,i}}}{\sum_{j=1}^{min(D,N)}{\sigma_{j,j}}}.
\end{align}
Concretely, the number of basis vectors $p$ is set at the maximum value, while ensuring that the contribution ratio does not exceed a threshold.
In the experiments, the threshold is selected by the grid-search algorithm. 

\subsection{Other Details}
All the experiments were conducted using computation servers equipped with NVIDIA's GeForce GTX 1080Ti and Tesla A100.
We used PyTorch~\cite{pytorch} to implement the experiments.

\section{Additional Experimental Results}

\subsection{Ablation Studies}
\subsubsection{Comparison with an Object-Aware Baseline}
We evaluated an another object-aware baseline regarding~\cite{objaware}, as shown in Table~\ref{tab:results_tab2}.
This baseline uses the sum of the original BLIP2s' score and the highest similarity between text and object embeddings (OEs) for retrieval.
OEs are obtained by applying ROI pooling to the visual tokens from the BLIP2s' backbone and parsing the pooled tokens to Q-Former.
We could see the effectiveness of our method again, as our method outperforms this simple baseline.

\subsubsection{Performance Analysis of Query Enhancement}
We evaluated the modified Q-Pert. (OrQ-Pert.) to confirm the effectivenss of our method.
Orthogonal Q-pert. (OrQ-Pert.) uses the projected query into an orthogonal complement of K-subspace instead of our perturbed query, i.e., it reduces object awareness.
As shown in Table~\ref{tab:results_tab2}, OrQ-Pert. shows lower performances, especially on small objects.
The result supports the effectiveness of our method.

\input{supp_tab/eccv_supp_ablation}

\subsection{Computational Cost}
Table~\ref{tab:cost} shows the extra computational cost.
The extra costs are not dominant.
The main bottleneck is the visual backbone, which is also the same as COCA and InternVL.
It would be an excellent future direction to consider an algorithm for pruning target cross-attention layers of our method to reduce the extra cost further.
Note that image features can be pre-computed for text-to-image retrieval.

\input{supp_tab/computational_time}

\subsection{Sensitivity on Hyper Parameters}
\input{supp_tab/hypara}
Table~\ref{tab:scale_sensitivity}~\ref{tab:sdim_sensitivity} shows retrieval performances by our method with varying hyperparameters.
We can see that the proposed method has low sensitivities to the hyperparameters.

\subsection{Additional Examples}

\begin{figure}[t]
    \centering
    \includegraphics[width=0.85\columnwidth]{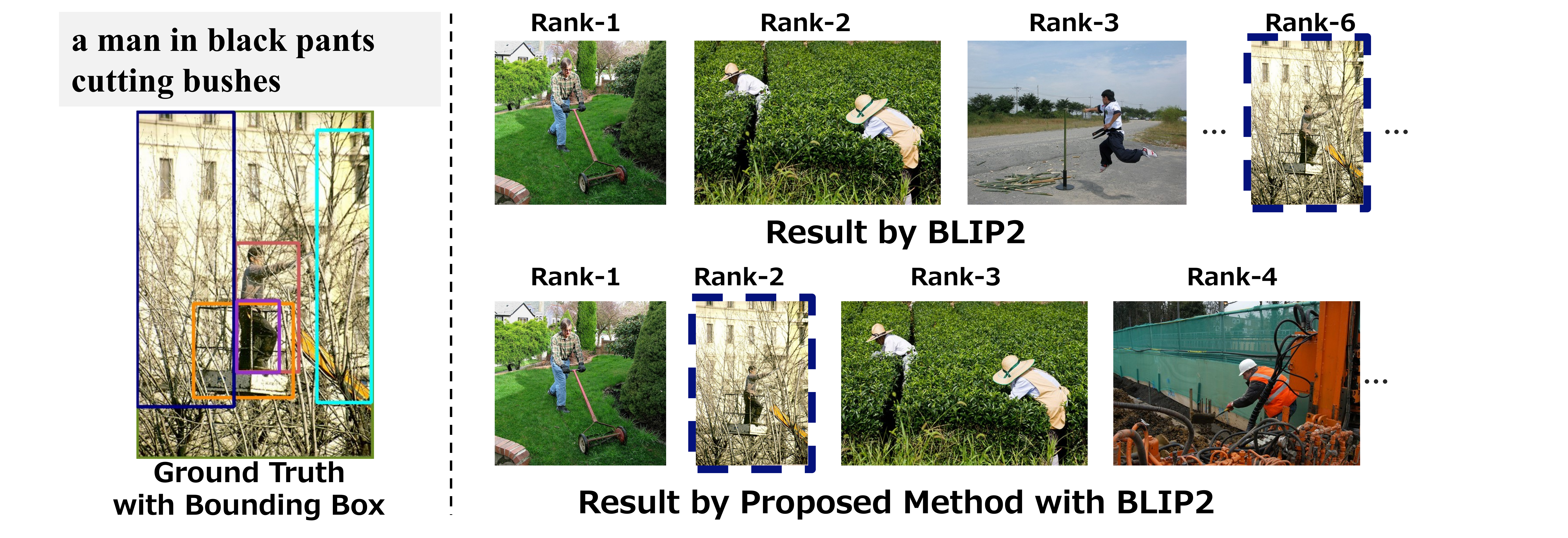}\\
    \hspace{-5mm}\includegraphics[width=0.85\columnwidth]{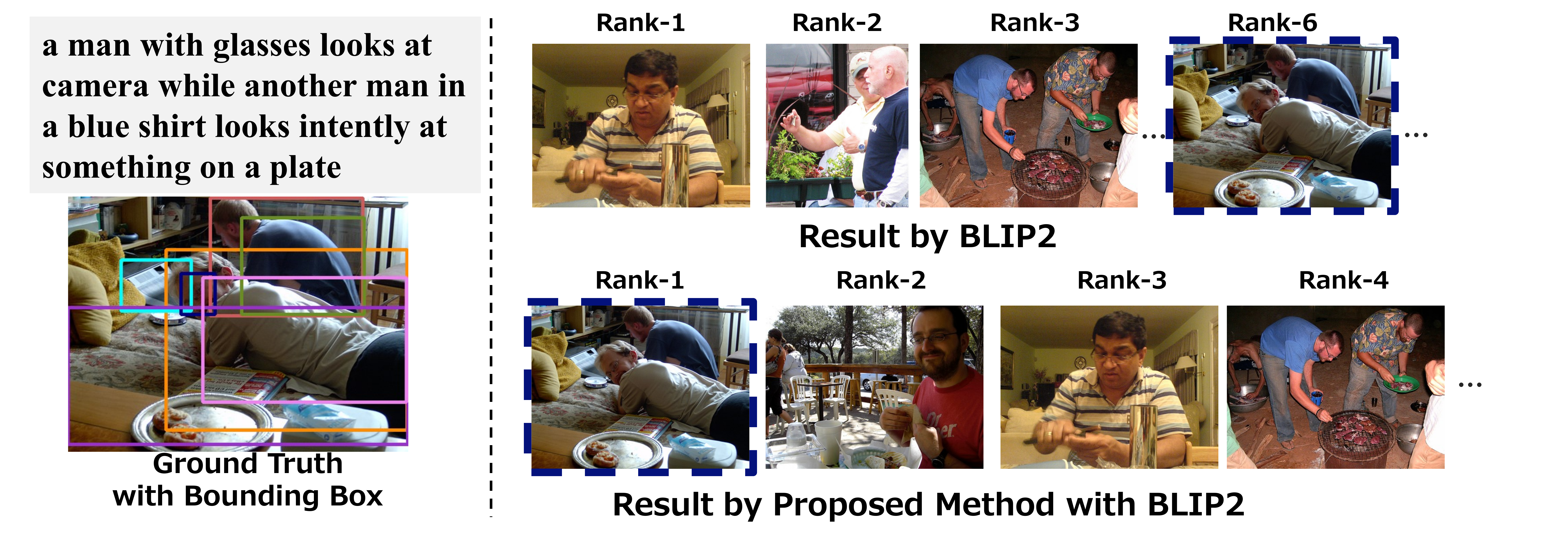}\\
    \includegraphics[width=0.85\columnwidth]{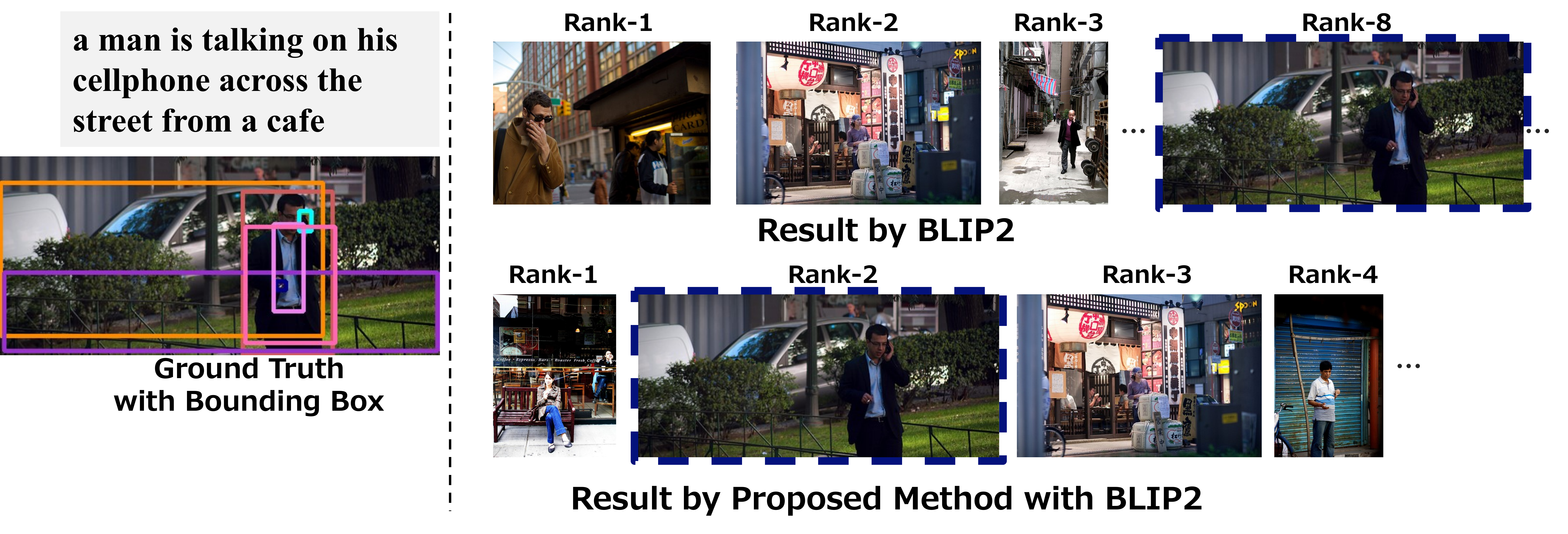}\\
    \includegraphics[width=0.85\columnwidth]{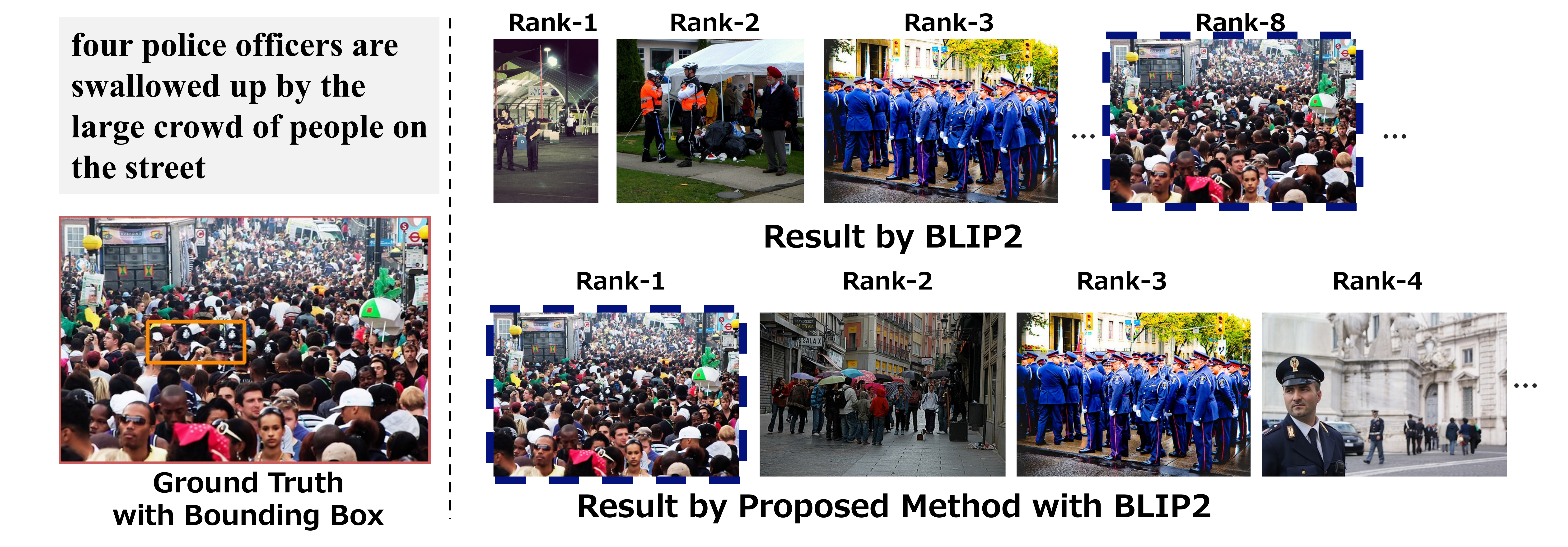}\\
    \includegraphics[width=0.85\columnwidth]{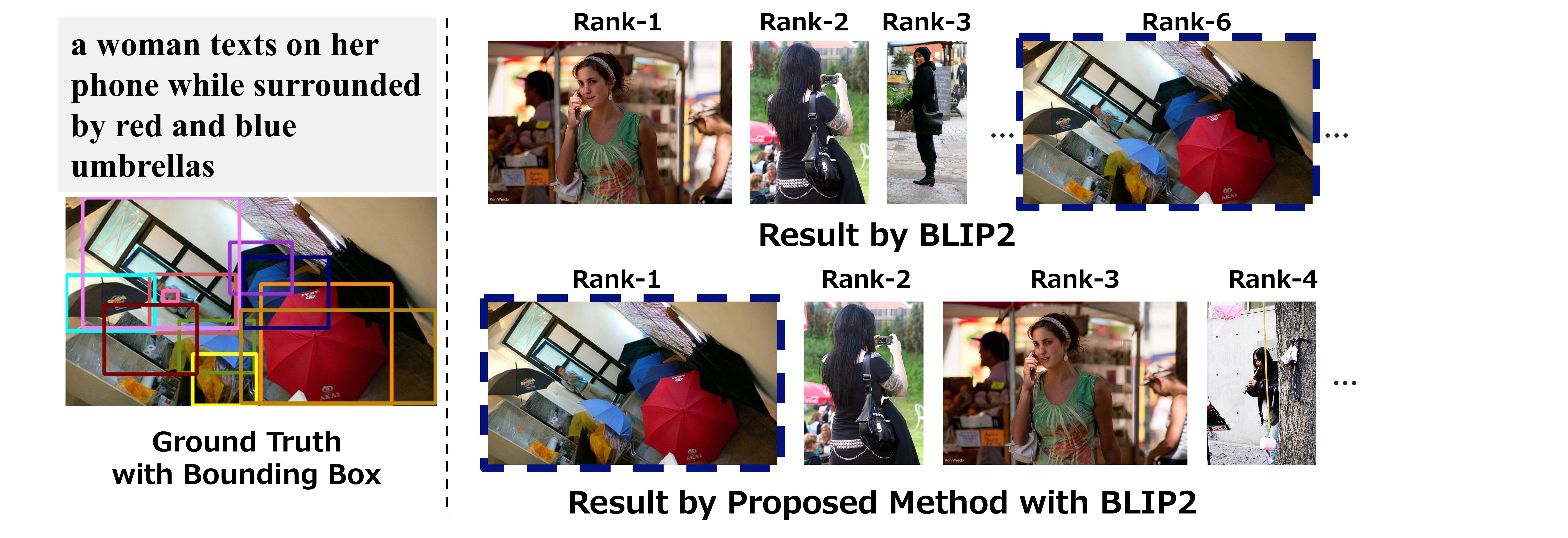}\\
    %\vspace{-0.2cm}
    \caption{Examples of image retrieval results by BLIP2~\cite{BLIP2BootstrappingLanguageImagea} and our method.}
    
    \label{fig:good_examples}
    %\vspace{-0.2cm}
\end{figure}

Figure~\ref{fig:good_examples} shows examples of image retrieval by BLIP2 and BLIP2 with Q-Perturbation.
It seems that our method could perform detailed matching between a text and images, making accurate retrievals.

\begin{figure}[t]
    \centering
    \includegraphics[width=0.85\columnwidth]{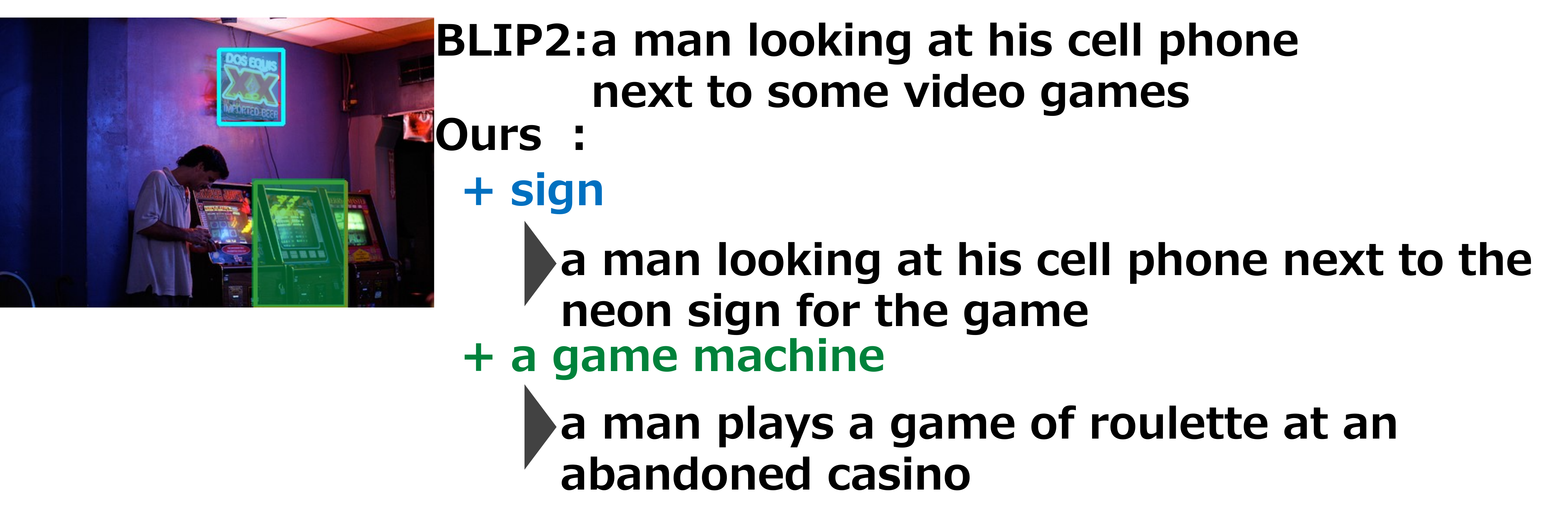}\\
    \includegraphics[width=0.85\columnwidth]{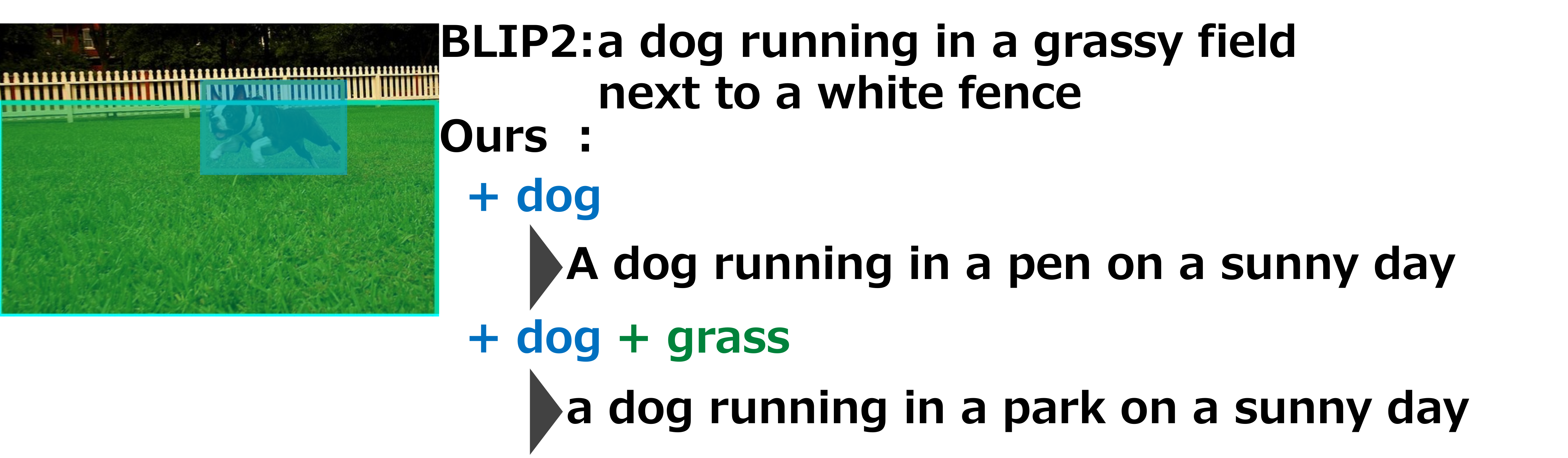}\\
    \includegraphics[width=0.85\columnwidth]{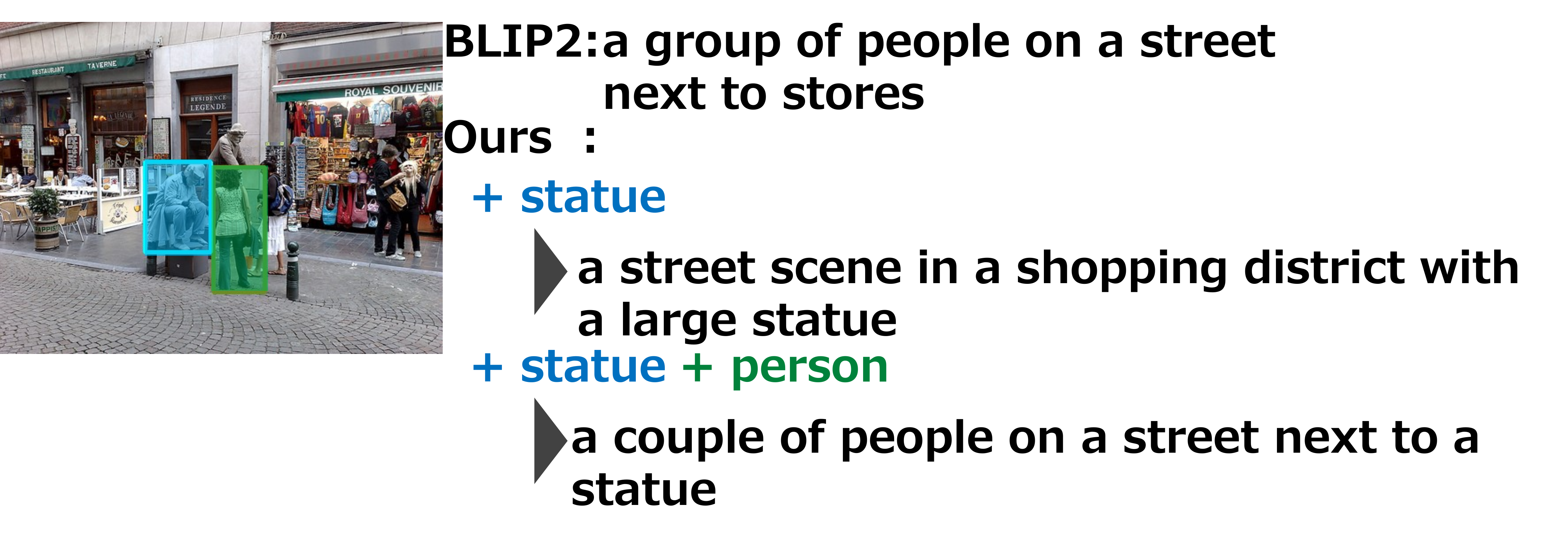}\\
    \includegraphics[width=0.85\columnwidth]{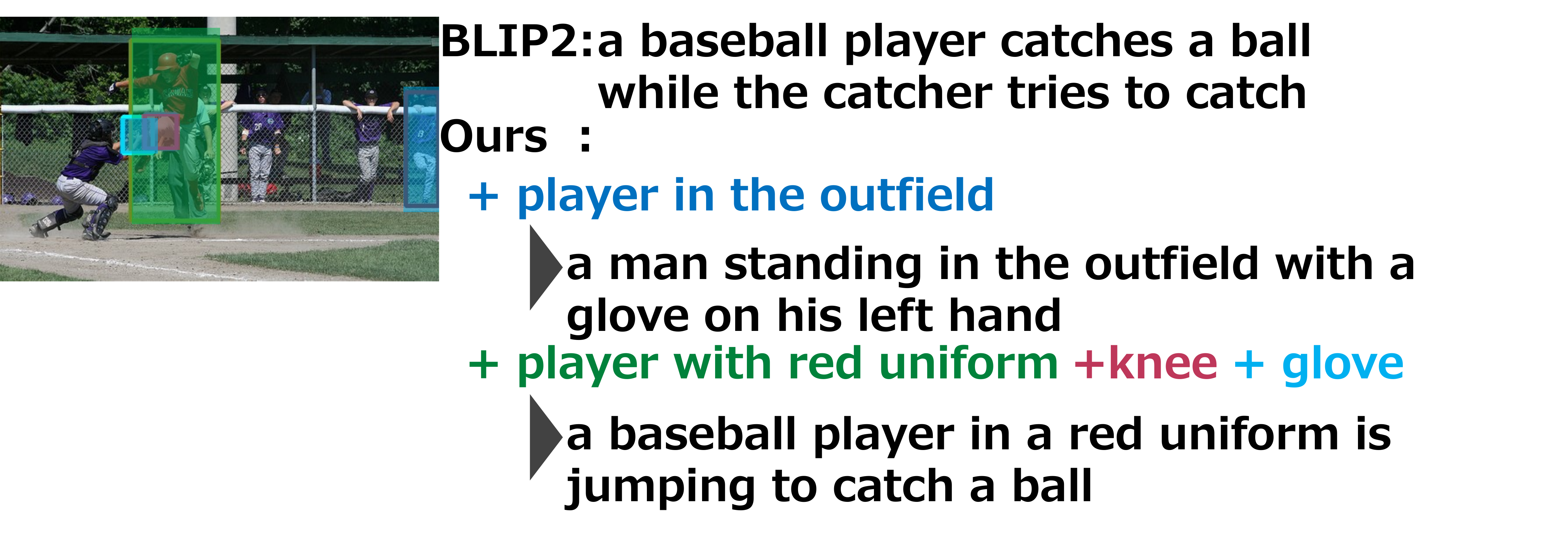}\\
    \caption{Image captioning results by BLIP2 and our method. ``+ word'' means enhancing information of the corresponding object}
    
    \label{fig:cap_examples}
    %\vspace{-0.2cm}
\end{figure}

Figure~\ref{fig:cap_examples} shows examples of image captioning results by BLIP2 and BLIP2 with Q-Perturbation.
The results suggests that our method enables captioning by a V\&L model to be controllable using object information.

%% file: supp_tab/psudo_code.tex
\begin{figure}[t]
\begin{lstlisting}[language=Python, label=code1, style=myPython, caption=PyTorch-like pseudo code for our Q-Perturbation.]
    
def query_perturbation(
    q: torch.Tensor,
    k: torch.Tensor,
    obj_mask: torch.Tensor,
    alpha: torch.Tensor,
    contrib_th: float,
):
    """Query-Perturbation.

    Apply query perturbation to the queries using the keys and object masks.

    Args:
        q (torch.Tensor): Queries. Shape: [#queries, dims]
        k (torch.Tensor): Keys. Shape: [#keys, dims]
        obj_mask (torch.Tensor): Object masks. Shape: [#objects, #keys]
        alpha (torch.Tensor): Weight scale. Shape: [#objects]
        contrib_th (float): Threshold for PCA.

    Returns:
        torch.Tensor: Perturbed queries. Shape: [#queries, dims]
    """
    pert_vec = torch.zeros_like(q)
    for i in range(obj_mask.shape[0]):
        # 1. Extract the keys corresponding to the object
        k_obj = k * obj_mask[i].unsqueeze(1)

        # 2. Make the key subspace; Apply PCA to the keys
        phi = PCA(k_obj, contrib_th)  # phi = [#n_components, dims]
        proj_mat = phi.T @ phi  # proj_mat = [dims, dims]

        # 3. Decompose the queries with the key subspace
        q_proj = q @ proj_mat  # q_proj = [#queries, dims]
        pert_vec += alpha[i] * q_proj

    # 4. perturb the queries
    return q + pert_vec
\end{lstlisting}
\end{figure}

%% file: supp_tab/eccv_supp_ablation.tex
\begin{table}[t]
    \caption{{\small Comparative results on Flickr30K for the T2I task  with two baselines, OEs and OrQ-Pert.}}
    \label{tab:results_tab2}
    \centering
        \begin{tabular}{lrrrrr}
            \toprule
                 &   ~10\%& R@1&  R@5&  mR@1&  mR@5 \\
            \midrule
            BLIP2         &  81.33&89.76 & {98.18} & 88.75 & 97.85  \\
            {\bf~w/Q-Pert. (E)}&  {\bf 84.00}&{\bf 89.82}& {\bf 98.20}&  {\bf 89.10}& {\bf 97.89} \\
\hdashline
            ~w/OEs &  56.00&78.48& 94.84& 74.28& 94.10 \\
            ~w/OrQ-Pert. &  77.33& {89.64}& {98.12}&  {88.16}& {97.53} \\
            \bottomrule
        \end{tabular}
\end{table}

%% file: supp_tab/computational_time.tex
\begin{table}
    \centering
    \caption{{\small Computational time [milliseconds] of visual feature extraction by VLMs. Q-Pert.(KC) and (QE) show the extra costs by K-subspace Construction and Query-Enhancement.}}  
    \begin{tabular}{lrrrr}
    \toprule
           GPU &  BLIP2 &  w/Q-Pert. &  Q-Pert.(KC)& Q-Pert.(QE)\\
    \cmidrule(lr){1-1}\cmidrule(lr){2-3}\cmidrule(lr){4-5}
       GTX 1080Ti  & 321.0& 413.0& +74.9& +1.3\\
       RTX 8000 & 128.0 & 187.0 & +57.1 & +0.6 \\
    \bottomrule    
    \end{tabular}
    \label{tab:cost}
\end{table}

%% file: supp_tab/hypara.tex
\begin{table}
    \caption{Sensitivity on hyperparameters, i.e., the weight function, and scale factor, with two evaluation indexes.}
    \begin{minipage}[c]{.5\hsize}
        \centering
        \subcaption{Flickr-30K dataset. R@1 small(10\%).}
        \begin{tabular}{llrrrrr}
        \toprule
            && \multicolumn{5}{c}{scale}\\
        \midrule
            &&  2 &  4&  6&  8& 10\\
        \midrule
            &$\bar{S}_b$ &  81.33&  \second{82.67}&  81.33&  80.00& 81.33
\\
            &1-$\bar{S}_b$ &  \second{82.67}&  \second{82.67}&  80.00&  81.33& \first{84.00}
\\
            &$\bar{S}_b$-0.5 &  81.33&  81.33&  81.33&  \second{82.67}& \second{{82.67}}
\\
            &0.5-$\bar{S}_b$&  81.33&  \second{82.67}&  81.33&  81.33& 81.33
\\
            &constant&  {\second{82.67}}&  81.33&  81.33&  81.33& \first{84.00}
\\
        \bottomrule
        \end{tabular}
    \end{minipage}
    \begin{minipage}[c]{.5\hsize}
        \centering
        \subcaption{Flickr-30K dataset. R@1.}
        \begin{tabular}{llrrrrr}
        \toprule
            && \multicolumn{5}{c}{scale}\\
        \midrule
            &&  2 &  4&  6&  8& 10\\
        \midrule
            &$\bar{S}_b$ &  88.73&  88.65&  88.67&  88.67& 88.77
\\
            &1-$\bar{S}_b$ &  88.85&  88.63&  88.63&  \second{88.93}& 88.62
\\
            &$\bar{S}_b$-0.5 &  88.69&  88.41&  88.75&  88.46& 88.52
\\
            &0.5-$\bar{S}_b$&  88.53&  88.80&  88.81&  88.72& 88.63
\\
            &constant&  88.63&  \first{88.95}&  \second{88.93}&  88.67& 88.76
        \\
        \bottomrule
        \end{tabular}
    \end{minipage}
    \label{tab:scale_sensitivity}
\end{table}

\begin{table}
    \caption{Sensitivity on hyperparameters, i.e., the weight function, and dimension of subspace, with two evaluation indexes.}
    \begin{minipage}[c]{.5\hsize}
        \centering
        \subcaption{Flickr-30K dataset. R@1 small(10\%).}
        \begin{tabular}{llrrrrr}
        \toprule
            && \multicolumn{5}{c}{Contribution ratio
}\\
        \midrule
            &&  0.80&  0.85&  0.90&  0.95& 0.99
\\
        \midrule
            &$\bar{S}_b$ &  81.33&  81.33&  81.33&  81.33& 
82.67
\\
            &1-$\bar{S}_b$ &  81.33&  \second{84.00}&  \second{84.00}&  82.67& 
81.33
\\
            &$\bar{S}_b$-0.5 &  81.33&  80.00&  80.00&  80.00& 
81.33
\\
            &0.5-$\bar{S}_b$&  81.33&  81.33&  80.00&  81.33& 
81.33
\\
            &constant&  81.33&  81.33&  \first{85.33}&  82.67& \second{84.00}
\\
        \bottomrule
        \end{tabular}
    \end{minipage}
    \begin{minipage}[c]{.5\hsize}
        \centering
        \subcaption{Flickr-30K dataset. R@1.}
        \begin{tabular}{llrrrrr}
        \toprule
            && \multicolumn{5}{c}{Contribution ratio
}\\
        \midrule
            &&  0.80&  0.85&  0.90&  0.95& 0.99
\\
        \midrule
            &$\bar{S}_b$ &  88.60&  88.60&  88.75&  88.59& 
88.77
\\
            &1-$\bar{S}_b$ &  88.84&  \first{89.16}&  88.84&  88.81& 
88.62
\\
            &$\bar{S}_b$-0.5 &  88.83&  88.46&  88.43&  88.49& 
88.52
\\
            &0.5-$\bar{S}_b$&  88.78&  88.75&  88.51&  88.67& 
88.63
\\
            &constant&  88.57&  88.67&  \second{89.14}&  88.70& 88.76
\\
        \bottomrule
        \end{tabular}
    \end{minipage}
    \label{tab:sdim_sensitivity}
\end{table}